\pdfoutput=1
\documentclass[11pt]{article}

\usepackage[final]{acl}

\usepackage{times}
\usepackage{latexsym}

\usepackage[T1]{fontenc}
\usepackage[utf8]{inputenc}

\usepackage{microtype}

\usepackage{inconsolata}

\usepackage{graphicx}
\usepackage{enumitem}
\usepackage{booktabs}  
\usepackage{multirow}
\usepackage{pifont}
\usepackage{xspace}
\usepackage{listings}
\usepackage{xcolor}
\usepackage{colortbl}
\usepackage{nicematrix}
\usepackage{afterpage}
\usepackage{textcomp}

\definecolor{LightCyan}{rgb}{0.88,1,1}
\newcolumntype{a}{>{\columncolor{LightCyan}}c}
\definecolor{myred}{rgb}{0.6, 0, 0}

\newcommand{\cmark}{\ding{51}}
\newcommand{\xmark}{\ding{55}}
\newcommand{\ours}{HiCUPID\xspace}
\newcommand\jm[1]{\textcolor{black}{{#1}}}
\newcommand\xx[1]{\textcolor{black}{{#1}}}
\newcommand\final[1]{\textcolor{black}{{#1}}}
\title{Exploring the Potential of LLMs as Personalized Assistants:\\ Dataset, Evaluation, and Analysis}
\author{Jisoo Mok\textsuperscript{\textnormal{1}}\thanks{~~Equal Contribution (email: magicshop1118@snu.ac.kr)}, Ik-hwan Kim\textsuperscript{\textnormal{1*}}, Sangkwon Park\textsuperscript{\textnormal{1*}}, Sungroh Yoon\textsuperscript{\textnormal{1,2,3}}\thanks{~~Corresponding Author (email: sryoon@snu.ac.kr)} \\ \\
\textsuperscript{\textnormal{1}}Department of Electrical and Computer Engineering, Seoul National University \\
\textsuperscript{\textnormal{2}}Interdisciplinary Program in Artificial Intelligence, Seoul National University \\
\textsuperscript{\textnormal{3}}AIIS, ASRI, INMC, and ISRC, Seoul National University}

\begin{document}
\maketitle
\begin{abstract}
Personalized AI assistants, a hallmark of the human-like capabilities of Large Language Models (LLMs), are a challenging application that intertwines multiple problems in LLM research.
Despite the growing interest in the development of personalized assistants, the lack of an open-source conversational dataset tailored for personalization remains a significant obstacle for researchers in the field.
To address this research gap, we introduce HiCUPID, a new benchmark to probe and unleash the potential of LLMs to deliver personalized responses.
Alongside a conversational dataset, HiCUPID provides a Llama-3.2-based automated evaluation model whose assessment closely mirrors human preferences.
\final{We release our dataset, evaluation model, and code at \url{https://github.com/12kimih/HiCUPID}.}
% \final{The dataset and the evaluation model are released in our Hugging Face repository: \url{https://huggingface.co/12kimih}.
% The code to reproduce our results is published in: \url{https://github.com/12kimih/HiCUPID}.}
% As LLMs continue to be integrated into various facets of human lives, Link to dataset: \url{https://huggingface.co/arranonymsub} Anonymous to code: \url{https://github.com/anonymsubmission-arr/hi_cupid.git}
\end{abstract}

\section{Introduction}\label{sec:intro}
Large Language Models (LLMs)~\cite{gpt, llama} with near-human capability revolutionized data-driven Natural Language Processing (NLP).
The notable examples of real-world applications enabled by the introduction of LLMs include LLM-backed coding agents (\textit{e.g.,} Github Copilot\footnote{\url{https://copilot.cloud.microsoft/}}), creative writing (\textit{e.g.,} Notion AI\footnote{\url{https://www.notion.so/product/ai}}), and chatbots and assistants (\textit{e.g.,} ChatGPT\footnote{\url{https://chatgpt.com/}} and Claude\footnote{\url{https://claude.ai/}}).
As LLMs continue to be integrated into various aspects of human lives, personalizing the LLM's responses to a human user emerges as a natural next step in LLM research.
% QuillBot\footnote{\url{https://quillbot.com/}} 

While personalization has traditionally been studied within few specialized domains~\cite{kar2020multi, christakopoulou2023large}, harnessing the emergent capabilities of LLMs for personalization opens the door to new possibilities.
In particular, developing an LLM-powered personalized assistant is garnering attention as an exciting and complex application that spans several research problems, which we categorize into~\textbf{5 desiderata of a personalized AI assistant}: Adherence to User Information (AUI), Understanding of Implicit Information (UII), Reasoning from Multiple Information (MI), Long-context Modeling Capacity (LC), and Proactiveness of Responses (PR).
The definition of each desideratum is provided in Section~\ref{sec:proposed}.
% Personalization holds the key to providing an engaging user experience in diverse areas, such as a recommender system, healthcare, and virtual assistant.
% . and generate a tailored response that stays faithful to the dialogue history, no matter the length and expansiveness of the history.
% a personalized experience, involves understanding the user's ''persona," which refers to [], and generating a customized response that properly reflects.
% To truly function as a personalized assistant and provide customized user experience, the backbone LLM must be able to tailor its response to reflect the user’s unique experience and preferences. 
% Personalized assistant presents unique challenges, such as long-context following, multi-hop reasoning, and information retrieval.
% While personalization has been mostly studied within the scope of recommendation systems, 
% Personalization is a widely-studied topic within the scope of recommendation systems, etc. 

Despite the central role of personalization in building a helpful and engaging assistant, a proper public benchmark to train and evaluate LLMs as a personalized assistant is missing.
The LLM research has been dominated by the ``one-size-fits-all'' paradigm, which prioritizes the versatility of LLMs over their functionality in specific use cases~\cite{lamp}.
This emphasis on generalization propelled a release of numerous general-purpose datasets~\cite{hendrycks2020measuring, kwiatkowski2019natural, zellers2019hellaswag}.
In contrast, existing datasets for personalization are mostly constrained to the task of personalized text classification, which is inapt for assessing the personalized generation capability of LLMs.
Although some of them~\cite{lamp, pchatbot} are designed to study text generation ability, they do not satisfy the aforementioned desiderata.
% Related works on datasets and benchmarks for personalization are extensively discussed in Section~\ref{sec:related}.
% Although the recently-released LaMP~\cite{lamp} dataset includes an array of text classification and generation tasks that require personalization, none of them concerns with personalized dialogue agents.
% PChatbot~\cite{pchatbot}, a Chinese dataset, shares a similar goal.
% While ConvAI2 and LaMP ConvAI2 and LaMP are restrictive.
% Due to this emphasis on the generalization capability of LLMs, 

To address this critical research gap and facilitate future efforts toward building LLM-powered personalized assistants, we introduce ``\ours (\textbf{C}onversations with \textbf{U}ser \textbf{P}ersonal \textbf{I}nformation \textbf{D}ataset),'' a new synthetic, GPT-4o-generated dataset that incorporates the multi-faceted challenges of personalized AI assistants.
In~\tablename~\ref{table:desiderata}, we compare~\ours against existing datasets and benchmarks to showcase the advantages of \ours.
\jm{Each user in~\ours is defined with 25 personas, each one of which represents a distinct dimension of their character, a profile, which provides objective information about the user, and 10 personal schedules.
The user's personal information - personas, profile, and schedules - is revealed naturally throughout the dialogue history between the user and the assistant.
\ours provides single-info question-answer (QA) pairs to examine whether the LLM has identified the corresponding personal information from the dialogue history and multi-info QA pairs that require combining a persona and a profile to be answered.}
% (\textit{i.e.,} age, gender, personality, occupation, and income range)
% The dialogue history additionally contains cues to the user's 10 personal schedules to simulate how the user might interact with the assistant to organize his or her life.
% with each dialogue representing a distinct persona.
% The resulting 25 dialogues are aggregated into a single dialogue history of the user.
%that incorporate clues about them, and the  
% The understanding and incorporation of each persona is investigated through associated question-answer (QA) pair
% \ours consists of GPT-4o-generated dialogue history with 25 different types of implicit persona and associated question-answer (QA) pairs to test the understanding and incorporation of each persona.
% The discussion of potential research topics enabled by the release of this dataset.
% Will encourage future research. 
% an abbreviated overview of existing datasets to clarify their limitations.

In addition,~\ours sets forth targeted evaluation protocols for personalized assistants: GPT-4o-based human preference estimation and Llama-3.2-based automated evaluation.
Conventionally, BLEU~\cite{papineni2002bleu} and ROUGE-L~\cite{lin2004rouge} are adopted to evaluate LLMs' responses, although they were not devised to assess the conversational capability of LLMs.
Through the GPT-4o evaluation, we gather human-aligned evaluation results, which are then distilled into the Llama-3.2-3B model to obtain an off-the-shelf proxy evaluator.
Evaluation protocols of~\ours explicitly check whether model responses are personalized, yielding evaluation scores that exhibit a high correlation with human preferences.
% \jm{Placeholder: By demonstrating that they lack correlation with human preferences, we confirm that Rouge-L and BLEU scores are inappropriate for quantifying the quality of personalized dialogue responses.}
% because they fail to consider personalization.

With~\ours, we conduct extensive experiments to investigate the personalization ability of state-of-the-art closed- and open-source LLMs, in conjunction with four inference-time and three train-time popularly used LLM customization methods.
% key findings?
% (\textit{i.e.,} zero- and few-shot prompting, BM25- and Contriever-based retrieval-augmented generation (RAG)~\cite{gao2023retrieval}) 
% (\textit{i.e.,} Supervised Fine-tuning (SFT) and Direct Policy Optimization (DPO)~\cite{dpo}) 
% At the inference level, zero- and few-shot prompting and retrieval-augmented generation (RAG)~\cite{gao2023retrieval} are used to induce personalization.
% At the training-level, we utilize low-rank adaptation (LoRA)~\cite{hulora} and direct policy optimization (DPO)~\cite{dpo} to fine-tune the backbone LLM.
Our contributions are as follows:
\begin{itemize}
  \setlength\itemsep{0em}
\item{We introduce~\ours, a new benchmark for training and evaluating LLMs as personalized assistants.~\ours properly reflects the challenges that arise in an LLM-powered personalized assistant system.} % }
\item{In~\ours, we provide a Llama-3.2-based proxy evaluation model for the automated evaluation of generated responses. By setting the degree of personalization in model-generated responses as a main evaluation criterion, our proxy evaluator yields a metric that is well-aligned with human preferences.}
\item{Our extensive empirical results and analyses uncover the limitations and potential of LLMs as personalized assistants. The failure of popular approaches to personalization of LLMs confirms that~\ours is a challenging benchmark to probe the quality of LLM-powered personalized assistants.} 
\end{itemize}

\begin{table}[t]
\centering
% \footnotesize
% \setlength{\tabcolsep}{4pt}
% \renewcommand{\arraystretch}{1.1}
{\resizebox{1\columnwidth}{!}
{\begin{tabular}{l|ccccc}
\toprule
Dataset & AUI & UII & MI & LC & PR   \\
\bottomrule\toprule
PChatbot & \cmark & \cmark & \xmark & \xmark & \xmark \\
PersonaChat \& ConvAI2 & \xmark & \xmark & \xmark & \xmark & \xmark \\
PersonalityEDIT & \xmark & \xmark & \xmark & \xmark & \cmark\\
LaMP & \cmark & \cmark & \xmark & \xmark & \xmark \\
HiCUPID (Ours) & \textcolor{green}{\cmark} & \textcolor{green}{\cmark} & \textcolor{green}{\cmark} & \textcolor{green}{\cmark} & \textcolor{green}{\cmark} \\
\bottomrule
\end{tabular}}
}
% \vspace{-0.3em}
\caption{Unlike existing datasets for personalized text generation,~\ours reflects all desiderata of a personalized virtual assistant.}
% \vspace{-1.0em}
\label{table:desiderata}
\end{table}
% How~\ours distinguishes itself from existing datasets for personalized text generation.
\section{Existing Methods and Benchmarks for LLM Personalization Research}\label{sec:related}
% \subsection{Existing Methods of LLM Personalization}
% The objective of LLM personalization~\cite{tan2023user} is to tailor the LLMs' generation results to reflect a user's needs, preferences, and characteristics, all of which construct his or her persona.
Personalization of language models~\cite{tan2023user} is studied across various tasks, such as recommendation systems~\cite{kang2023llms}, long-form text generation~\cite{li2023teach}, proactive dialog systems~\cite{yang2021improving, shi2021refine}, and virtual assistants~\cite{mysore2023pearl, zhang2024llm}.
Existing endeavors toward personalization can largely be categorized into inference-time~\cite{dai2023uncovering, kang2023llms, wang2023learning, mysore2023pearl, lamp, salemi2024ragpeft, salemi2024optimization, richardson2023integrating, liu2024once} and train-time approaches~\cite{oppu, salemi2024ragpeft, li2023steerability, tang2023enhancing}
\jm{Due to the page constraint, a detailed discussion on existing personalization approaches has been moved to Appendix.}

\begin{figure*}[t]
  \centering
%   \vspace{-0.2em}
  \includegraphics[width=\linewidth]{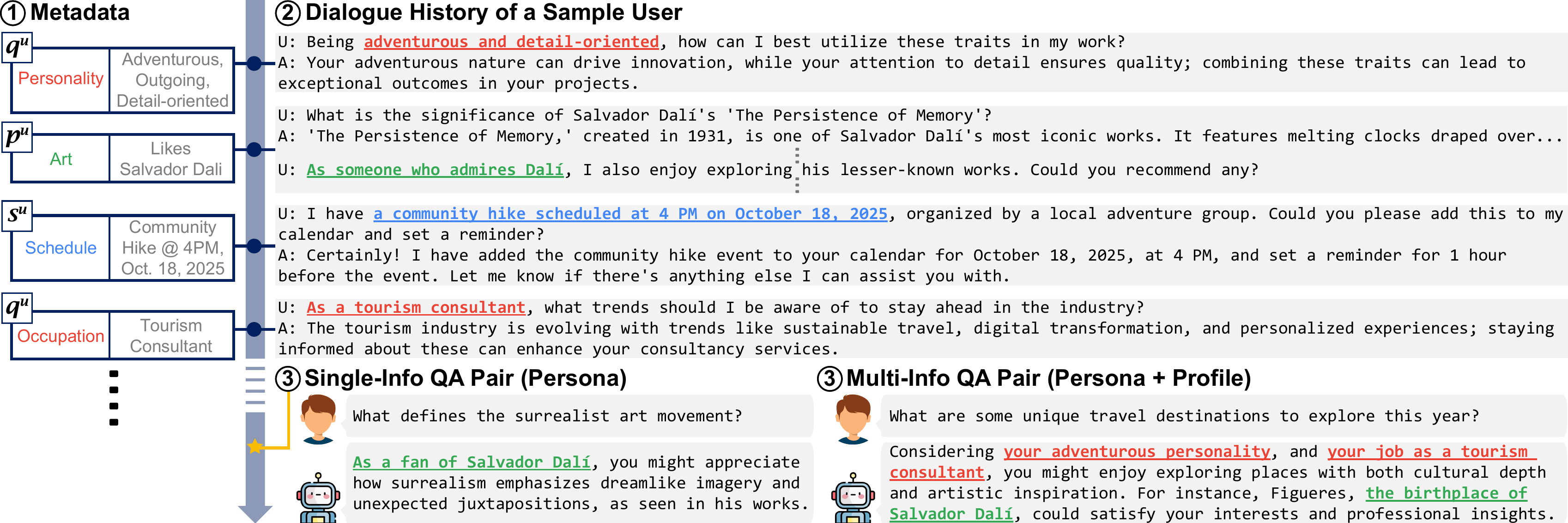}\\[-0.3em]
  \caption{\label{fig:our_framework} Configuration of~\ours. (1) A user $u$ is characterized by a set of metadata or personal information ($\mathcal{P}^u \cup \mathcal{Q}^u \cup \mathcal{S}^u$), which is only used for evaluation. (2) For each personal information, we create a dialogue that implicitly hints at it. All dialogues are aggregated into $\mathcal{D}^u$. (3) To probe whether the LLM picked up on one piece of personal information, we create a single-info QA pair for each persona or schedule. To study the LLM's multi-hop reasoning ability, we create multi-info QA pairs by pairing the user's profile with a closely-related persona.}
  \vspace{-5pt}
  % and must be excluded from the personalized generation process.
\end{figure*}
% Inference-time approaches either augment the input prompt with in-context demonstrations~\cite{dai2023uncovering, kang2023llms, wang2023learning} or employ retrieval-augmented generation (RAG) and its variants, such as profile-augmentation, to incorporate user-specific information~\cite{mysore2023pearl, lamp, salemi2024ragpeft, salemi2024optimization, richardson2023integrating, liu2024once}.
% Training-time methods~\cite{oppu, salemi2024ragpeft} rely on parameter-efficient fine-tuning (PEFT)~\cite{hulora, xu2023parameter}, or utilize data-driven methods to extract user's information more compactly to reduce noisy and fine-grained learning signals~\cite{li2023steerability, tang2023enhancing}.
% RAG-based methods address this limitation by retrieving the most relevant information from history.

% \subsection{Datasets and Benchmarks for LLM Personalization Research}\label{sec:2.2}
Despite the importance of personalization in real-world LLM applications, most of the existing datasets are restricted to a simple task of personalized text classification, rendering them inadequate for harnessing the emergent capabilities of LLMs for personalization.
In recommendation or review prediction tasks, personalization is studied with MovieLens~\cite{harper2015movielens}, Amazon~\cite{amazondataset}, or MIND~\cite{mindsmall} datasets, which are designed for text classification.
Personalized story or book evaluation studies, conducted on MPST~\cite{kar2020multi}, DOC~\cite{yang2023doc}, or Douban~\cite{zhugraphical} rating datasets, are classification tasks as well.
  
Although a limited number of datasets for personalized text generation exist, they lack the complexity to be used for developing a personalized assistant.
\jm{PersonaChat~\cite{personachat}, ConvAI2~\cite{convai2}, which extends PersonaChat, and PersonalityEDIT~\cite{personalityedit} are designed to equip LLMs with personality traits instead of personalizing their responses to a user.}
Therefore, their definition of ``personalization'' is inherently different from~\ours.
The LaMP~\cite{lamp} dataset provides an array of personalized text classification and generation tasks, but none of them is conversational.
Lastly, PChatbot~\cite{pchatbot} and Synthetic-Persona-Chat~\cite{jandaghi2023faithful} share the common goal of personalizing LLMs on user's traits, but their dialogues are shorter than the context length of current state-of-the-art LLMs.
% and thus cannot serve as a challenging benchmark.
% are too cursory and straightforward to be challenging and meaningful benchmarks for studying LLMs.
% PChatbot does not support multiple persona, long-context modeling, or proactiveness, nor does it release the persona descriptions as metadata for evaluation.}
% Likewise, PersonalityEDIT [does what].

\begin{table*}[t]
\centering
% \footnotesize
% \setlength{\tabcolsep}{4pt}
\renewcommand{\arraystretch}{0.9}
% {\resizebox{1\linewidth}{!}
% {\begin{NiceTabular}{l|c|c|c|c|c|c}
% \toprule
% Data Split & \# of Users & \# of Dialogues & \# of QA Pairs & Length of $d^u_i$ & Length of $\mathcal{D}^u$  & Length of a QA Pair \\
% \bottomrule\toprule
% Train Set & \multirow{2}{*}{1250} & \multirow{2}{*}{50000} & 40000 & \multirow{3}{*}{$371.7 \pm 245.1$
% } &  \multirow{3}{*}{$14868.3 \pm 543.7$} & \multirow{3}{*}{$77.1 \pm 28.1$} \\
% Test Set 1 & & & 10000 & & &  \\
% \cmidrule{1-4}
% Test Set 2 & 250 & 10000 & 10000 & & &  \\
% \bottomrule
% \end{NiceTabular}}
% }
{\resizebox{1\linewidth}{!}
{\begin{NiceTabular}{l|ccc|ccccc}
\toprule
\multirow{2}{*}{Data Split} & \multicolumn{3}{c}{Number} & \multicolumn{5}{c}{Length} \\
\cmidrule{2-9}
& Users & Dialogues & QA Pairs & $\mathcal{D}_p^u$ (Persona) & $\mathcal{D}_q^u$ (Profile) & $\mathcal{D}_s^u$ (Schedule) & $\mathcal{D}^u$ (Whole) & QA Pair  \\
\midrule
Train Set & \multirow{2}{*}{1250} &  \multirow{2}{*}{50000} & 40000 & \multirow{3}{*}{$15962.3 \pm 538.1$} & \multirow{3}{*}{$329.0 \pm 31.4$} & \multirow{3}{*}{$970.9 \pm 50.7$} & \multirow{3}{*}{$17256.3 \pm 543.7$} & \multirow{3}{*}{$57.3 \pm 17.9$} \\
Test Set 1 &  &  & 10000 &  &  &  &  &  \\
\cmidrule{1-4}
Test Set 2 & 250 & 10000 & 10000 &  &  &  &  &  \\
\bottomrule
\end{NiceTabular}}
}
\vspace{-0.3em}
\caption{\ours dataset statistics. $\mathcal{D}^u$ of a user $u$ consists of persona, profile, and schedule dialogues: $\mathcal{D}^u_p$, $\mathcal{D}^u_q$, and $\mathcal{D}^u_s$. The length of dialogues and QA pairs is quantified in the number of GPT-2 Tokenizer~\cite{radford2019language} tokens. Test 1 and Test 2 splits denote Seen User/Unseen QA pair and Unseen User/Unseen QA pair  settings.}
\vspace{-5pt}
\label{table:datastat}
\end{table*}

%% 여기 고쳐야함 %%
% Dialogue:
% Persona 548.4 +- 112.6 / 13709.3 +- 538.1
% Schedule 87.8 +-12.1 / 877.9 +- 50.7
% profile 56.2 +- 8.1 / 281.0 +- 31.4
% QA:
% 67.9 +- 23.9
\section{HiCUPID: Dataset Configuration and Generation}\label{sec:proposed} 
% \noindent\textbf{Problem Formulation} Given a sequence of input tokens $x$, LLMs generate the most probable next sequence of output tokens $y$.
% To generate a personalized response specific to a user $u$, LLMs' generation results must additionally be conditioned on a set of user's personas $\mathcal{P}^u$.
Below, we outline \textbf{5 desiderata} that LLMs must satisfy for them to be deployed seamlessly as personalized assistants. \\
% \noindent\textbf{5 Desiderata of Personalized Virtual Assistants}
% \begin{enumerate}[label=\alph*)]
\noindent(a)~\textit{Adherence to User Information (AUI)}: A personalized assistant must respond specifically in a user-aware manner. Thus, as discussed in Section~\ref{sec:related}, personalization in our work refers to conditioning LLM's responses on user's information, instead of assigning LLMs a personality.\\
\noindent(b)~\textit{Understanding of Implicit Information (UII)}: Explicitly supplying the LLM with a user's personal information is often infeasible because such information is fluid and evolves over time. In the absence of explicit cues, the LLM must infer relevant information from its interactions with the user.\\
\noindent(c)~\textit{Reasoning from Multiple Information (MI)}: Multiple pieces of personal information appear scattered throughout the dialogue history. Therefore, the LLM must be able to combine and reason from all of the extracted information to customize its response to the user.\\
\noindent(d)~\textit{Long-context Modeling Capacity (LC)}: As more exchanges between the user and the LLM occur, more personal information is gradually revealed, while the limited context length of LLMs makes it increasingly challenging for the LLM to integrate information from past interactions. Nonetheless, the LLM should be able to retain information from any point in the dialogue history.\\
% no matter how expansive the dialogue history is.\\
\noindent(e)~\textit{Proactiveness of Responses (PR)}: The response of a personalized assistant should not only adhere to the dialogue history, but it also needs to provide proactive recommendations or suggestions based on the user's persona.
% {If the LLM does no more than passively respond to the user's questions, is it really an assistant?} 
% \jm{Placeholder: The LLM response should reflect the user's personas. The LLM response should reflect the user's personas. The LLM response should reflect the user's personas.}
% To function as a personalized assistant, the LLM first needs to extract the persona information implicitly embedded within the dialogue history. 
% Afterward, the LLM must be able to perform personalization by combining fragments of persona information that are scattered throughout the dialogue history, n
% Lastly, not only should

The comparison of notable datasets used for personalized text generation in~\tablename~\ref{table:desiderata} shows that no existing dataset adequately reflects these challenges, presenting a serious roadblock in developing a personalized assistant.
To fill this major research gap, we introduce~\ours, a synthetic dataset that consists of dialogue history and QA pairs.
They are accompanied by the visual illustration of~\ours in Figure~\ref{fig:our_framework} and the summary of dataset statistics in~\tablename~\ref{table:datastat}.
% The design of~\ours incorporates all of the aforementioned desiderata of a personalized assistant.
The following sections detail how~\ours is constructed to test the ability of LLMs to generate personalized responses.
% However, when interacting with an assistant, a human user would not provide it with such explicit and highly-formatted personal information.
% Therefore, the data generation process, which inherently violates the five desiderata, fails to mimic realistic interactions between a user and an assistant.}
% While LLMs’ ability to execute personalization given restrictive assumptions allows the construction of the proposed dataset, the task of realistic and natural personalized response generation is yet to be fully resolved due to the impracticality of the aforementioned assumptions. 
%  of~\ours are summarized in~\tablename~\ref{table:desiderata}.
% The persona information  provided as metadata for evaluation purposes.

\subsection{User Metadata of~\ours}
% \jm{To create synthetic users, we first define each user's personal information or metadata 
\jm{Synthetic users in~\ours are defined with the following set of personal information: 25 personas, five pieces of profile information, and 10 schedules.} \\
\noindent\textbf{Persona:} The persona of a user $u$ is defined over 25 persona dimensions (\textit{e.g.,} Sports, Music, Fashion, etc.), \jm{a set of preferences, opinions, or experiences that shape the user.}
The comprehensive list of persona dimensions is in Section~\ref{sec:our_personas} of Appendix.
In each persona dimension, we define 150 distinct personas as combinations of a relation (\textit{e.g.,} likes/dislikes, supports/does not support, etc.) and an entity (\textit{e.g.,} Soccer, Baseball, etc. in the ``Sports'' dimension).
We sample one persona $p$ per persona dimension and assemble them into a set of user's personas $\mathcal{P}^u = \{p^u_1, p^u_2, ..., p^u_N\}$, where $N=25$. 
We assume that 10 users can have a common persona; for instance, it is reasonable to assume that 10 users simultaneously like Soccer.
Given this assumption on the overlap of personas among a small subset of users, we create 1,500 synthetic users.\\
% each with a unique set of personas.\\ 
\noindent\textbf{Profile:} \jm{contains five pieces of objective information about the user: age, gender, personality, occupation, and income range. 
To generate 1,500 synthetic profiles, we randomly sample 1,500 individuals from PersonaHub~\cite{billion_persona}, a collection of personas and characters curated from the web.
Then, GPT-4o is prompted to extrapolate the profile of each individual using the template in Figure~\ref{fig:p_profile_metadata} of Appendix.
Each profile $\mathcal{Q}^u = \{q^u_1, q^u_2, ..., q^u_M\}$, where $M=5$, is paired with a user $u$.} \\
\noindent\textbf{Schedule:} \jm{is comprised of an event or a task and a timestamp. 
10 schedules of a user $u$, $\mathcal{S}^u = \{s^u_1, s^u_2, ..., s^u_L\}$, where $L=10$, are generated by GPT-4o given the user's profile $\mathcal{Q}^u$ to ensure that they are realistic and feasible. The prompt template for schedule metadata generation can be found in Figure~\ref{fig:p_schedule_metadata} of Appendix.} 

The metadata of a user $u$ is constructed by combining all 25 personas, five pieces of profile information, and 10 schedules: $\mathcal{U}^u = \mathcal{P}^u \cup \mathcal{Q}^u \cup \mathcal{S}^u$.
These pre-defined metadata form the basis of the synthetic dialogues and QA pairs in~\ours.
In practice, the usage of metadata is strictly restrained to evaluation purposes to test the UII desideratum.
% [It should not be used to personalize]
% are used to construct a set user $u$.
% Consequently, a set of user's persona information 
% For each user $u$, we define a set of persona information $\mathcal{P}^u = \{p^u_1, p^u_2, ..., p^u_N\}$, where $N$ denotes the number of persona dimensions per user.
% We assume that 2 users can share a piece of persona information; for instance, it is reasonable to assume that 2 users both like Cricket.

\begin{figure*}[t]
  \centering
%   \vspace{-0.2em}
  \includegraphics[width=\linewidth]{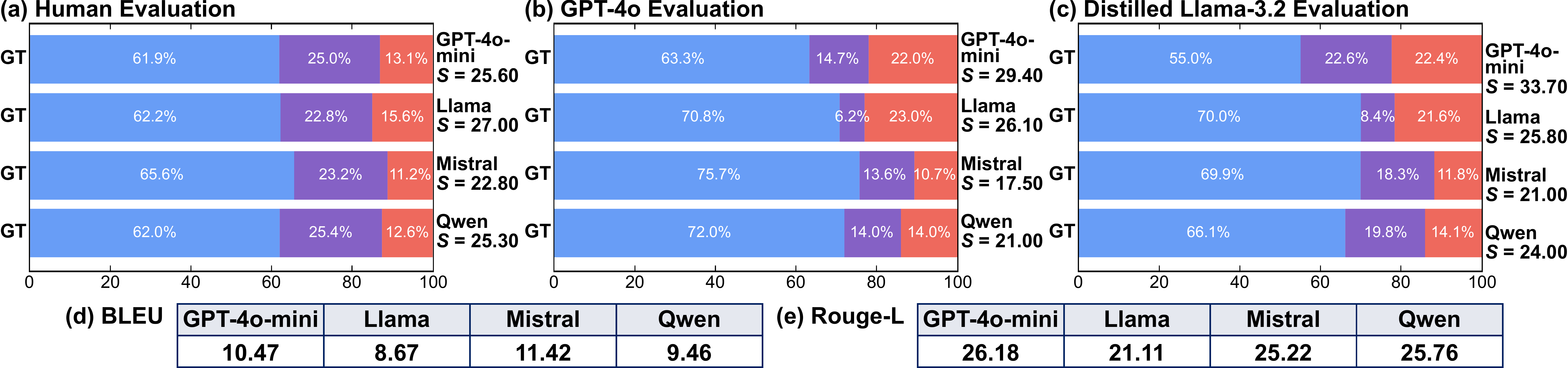} \\[-0.3em]
  \caption{\label{fig:human_gpt} Evaluation of 100 zero-shot model-generated responses with human evaluators~\textit{vs.} GPT-4o~\textit{vs.} Distilled Llama-3.2. Blue, Purple, and Red bars correspond to the GT Win, Tie, and Model Win Rates, respectively. }
  \vspace{-5pt}
\end{figure*}

\subsection{Dialogues}
\noindent\textbf{Persona:} For each user, we generate 25 persona dialogues, which correspond to 25 persona dimensions: $\mathcal{D}_p^u=\{d^u_{p_1}, ..., d^u_{p_{25}}\}$. Even if 10 users share a common persona, it is unnatural for them to have exactly the same conversation with an assistant.
Therefore, for a persona $p$, we generate 10 different versions of dialogues wherein the user provides the assistant with hints to $p$. The prompt template used to generate 10 distinct persona dialogues is given in Figure~\ref{fig:p_persona_dialogue} of Appendix.
The prompt enforces that the persona is revealed naturally amidst the dialogue, and that \jm{each persona dialogue is structured to contain 10 turns. 
% To guarantee that 10 diversified dialogues are generated with the same prompt, the GPT-4o API generation setting was set to $\tau=1.0$ and $p=0.95$, where $\tau$ and $p$ denote the temperature of a probability distribution and the nucleus sampling probability, respectively.
} \\
\noindent\textbf{Profile and Schedule:} \jm{Each user additionally comes with five profile dialogues, associated with five pieces of profile information, and 10 schedule dialogues: $\mathcal{D}_q^u=\{d^u_{q_1}, ..., d^u_{q_{5}}\}$ and $\mathcal{D}_s^u=\{d^u_{s_1}, ..., d^u_{s_{10}}\}$. Profile and schedule dialogues contain a single turn, with the user asking a question or making a request and the assistant responding to the user. The prompt templates used to generate profile- and schedule dialogues are provided in Figures~\ref{fig:p_profile_dialogue} and~\ref{fig:p_schedule_dialogue} of Appendix.
}

\jm{All three types of dialogues are aggregated into the dialogue history of a user: $\mathcal{D}^u = \mathcal{D}_p^u \cup \mathcal{D}_q^u \cup \mathcal{D}_s^u$.}
As reported in~\tablename~\ref{table:datastat}, the resulting dialogue history contains up to $17$k tokens on average, \jm{which is sufficiently extensive to test if the LLM's long-context handling ability meets the LC desideratum.}
% have with a virtual assistant. 
% hints to $p$.
% For each pre-defined persona information $p$, 
% To simulate a user's interaction, \ours largely consists of 

% \input{tex/prelim_bleu_rouge}
\subsection{Single- and Multi-Info QA Pairs}
\noindent\textbf{Single-Info QA:} Every persona and schedule dialogue comes with a QA pair designed to probe the LLM's awareness of the corresponding information.
Thus, each user has 35 single-info QA pairs that require only one persona or schedule to be considered when answering the question.\\
\noindent\textbf{Multi-Info QA:} To study the MI desideratum,~\ours provides five multi-info QA pairs, which need to be answered by combining one persona from $\mathcal{P}^u$ and the profile $\mathcal{Q}^u$.
To create multi-info QA pairs, we generate five realistic combinations of a persona and a profile with the prompt template in Figure~\ref{fig:p_2info_metadata} of Appendix. 

All QA pairs include a personalized and a general answer, which can be used as in-context demonstrations or positive/negative instances for reward modeling.
Prompt templates to generate QA pairs are in Figures~\ref{fig:p_persona_qa} (persona),~\ref{fig:p_schedule_qa} (schedule), and~\ref{fig:p_2info_qa} (persona+profile).
As in the persona dialogue generation process, the prompt template for persona QA pairs generates 10 different persona QA pairs for 10 users.
For the profile and schedule QA pair types, the prompt only generates one QA pair at a time since every user has a disparate set of schedules and persona-profile combinations.
Following the UII desideratum, the prompt precludes the user's question from explicitly referring to their personal information to maintain the implicitness of personal information.
% and then generate one QA pair per combination.

Note that the ability of GPT-4o to generate synthetic dialogues and QA pairs does not imply that GPT-4o addresses the challenges in personalized assistant development.
To create synthetic dialogues and QA pairs with GPT-4o, user's personal information is explicitly supplied within prompts, which are heavily-engineered through OpenAI's meta-prompt provided in Figure~\ref{fig:meta_p} of Appendix.
However, a realistic personalized assistant that is compliant with the five desiderata must be able to provide personalized responses even without explicit and highly-formatted personal information. 
% fully solves the problem of personalized response generation.
% However, such explicit and highly-formatted personal information may not always be available as elaborated in the UII desideratum of a personalized assistant.
% Section~\ref{sec:} when interacting with a user. 
% Therefore, the ability of GPT-4o 
% to generate synthetic dialogues and QA pairs cannot be equated to fully solving 
% However, at inference time, no such information is given to the model, requiring it to infer personal information from the preceding dialogues.

\jm{In summary,~\ours is configured to study whether LLMs can personalize its response given the dialogue history $\mathcal{D}^u$ while satisfying the five desiderata.
Section~\ref{sec:cupid_desiderata} of Appendix discusses how the design of~\ours probes of all five desiderata of a personalized assistant in further detail.} 
\ours offers two evaluation settings, the Seen User (Dialogue History) / Unseen QA Pair (Test Set 1) and the Unseen User (Dialogue History) / Unseen QA Pair test splits (Test Set 2), depending on whether the user's dialogue history is available at train time.
Among the 1,500 synthetic users, 250 are set aside for Test Set 2.
The QA pairs of the remaining 1,250 users are split with the ratio of $4:1$ to construct the Train Set and Test Set 1.
% \subsection{Train-Test Data Splits}
% \jm{Test 1 is intended for assessing generalization capability to unseen QAs, while Test 2 is intended for assessing it on completely unseen users.
% Intuitively, Test 2 is the more challenging test setting of the two.}

\begin{table*}[t]
\centering
\footnotesize
\resizebox{1\linewidth}{!}{
\begin{NiceTabular}{l|l|c|c|cccc|cccc}
\toprule
\multirow{2}{*}{Model} & \multirow{2}{*}{Method} & BLEU & ROUGE-L & \multicolumn{4}{c|}{GPT-4o Score ($S_\textrm{GPT}$)} & \multicolumn{4}{c}{Llama 3.2 Score ($S_\textrm{Llama}$)} \\
& & \multicolumn{1}{c}{(Total)} & \multicolumn{1}{c}{(Total)}  & \multicolumn{1}{c}{Persona} & \multicolumn{1}{c}{Schedule} &  \multicolumn{1}{c}{Multi-Info} & \multicolumn{1}{c}{Total} &  \multicolumn{1}{c}{Persona} & \multicolumn{1}{c}{Schedule} &  \multicolumn{1}{c}{Multi-Info} & \multicolumn{1}{c}{Total} \\
\bottomrule
\toprule
\multirow{2}{*}{GPT-4o-mini} & 0-shot & 8.2 & 19.7 & \textbf{42.1} & 9.5 & \textbf{4.4} & 28.0 & \textbf{44.7} & 8.8 & 10.8 & 30.4 \\
 & 3-shot & \textbf{16.1} & \textbf{30.0} & 40.5 & \textbf{76.1} & 4.2 & \textbf{35.3} & 42.6 & \textbf{75.4} & \textbf{11.4} & \textbf{37.5} \\ \midrule
\multirow{7}{*}{Llama-3.1-8B} & 0-shot & 8.2 & 19.2 & 38.0 & 13.9 & 3.5 & 25.9 & 39.7 & 9.4 & 8.1 & 27.0 \\
 & 3-shot & 16.3 & 29.6 & 39.4 & 49.8 & 6.3 & 31.6 & 38.8 & 48.3 & 12.3 & 31.8 \\
 & BM25 & 9.4 & 21.8 & 29.7 & 84.3 & 2.3 & 29.4 & 34.1 & 78.5 & 6.1 & 31.9 \\
 & Contriever & 9.5 & 22.1 & 38.8 & 75.4 & 4.2 & 34.2 & 42.6 & 70.3 & 9.8 & 36.6 \\
 & SFT & 24.9 & 38.5 & 36.2 & 88.0 & 12.4 & 35.2 & 36.5 & 87.5 & 15.8 & 35.7 \\
 & DPO & 7.6 & 17.9 & 24.8 & 4.9 & 2.1 & 16.4 & 34.8 & 4.2 & 6.4 & 23.1 \\ 
 & SFT$+$DPO & \textbf{27.6} & \textbf{42.7} & \textbf{49.1} & \textbf{98.6} & \textbf{14.5} & \textbf{44.8} & \textbf{48.1} & \textbf{98.1} & \textbf{18.4} & \textbf{44.6} \\ \midrule
\multirow{7}{*}{Mistral-7B} & 0-shot & 8.6 & 19.1 & 20.9 & 0.0 & 1.5 & 13.3 & 30.5 & 0.0 & 3.8 & 19.5 \\
 & 3-shot & 10.3 & 21.6 & 28.6 & 6.3 & 3.5 & 19.1 & 36.2 & 5.6 & 7.6 & 24.2 \\
 & BM25 & 7.3 & 18.0 & 41.0 & 8.6 & 4.9 & 27.3 & 43.7 & 6.3 & 9.0 & 29.2 \\
 & Contriever & 7.5 & 18.5 & \textbf{48.8} & 7.9 & 7.4 & 32.4 & \textbf{51.6} & 5.9 & 13.6 & 34.7 \\
 & SFT & 32.1 & 46.0 & 27.6 & \textbf{99.8} & 15.1 & 31.6 & 31.2 & \textbf{99.8} & 19.7 & 34.4 \\
 & DPO & 8.4 & 16.9 & 8.2 & 2.2 & 0.2 & 5.4 & 6.4 & 1.4 & 0.5 & 4.2 \\ 
 & SFT$+$DPO & \textbf{32.4} & \textbf{46.7} & 44.7 & 99.7 & \textbf{17.6} & \textbf{42.6} & 44.8 & \textbf{99.8} & \textbf{20.4} & \textbf{43.0} \\
 \midrule
\multirow{7}{*}{Qwen-2.5-7B} & 0-shot & 8.1 & 18.6 & 26.6 & 0.0 & 3.0 & 17 & 34.6 & 0.0 & 6.1 & 22.4 \\
 & 3-shot & 12.5 & 24.2 & 24.6 & 29.8 & 2.1 & 19.4 & 32.6 & 28.7 & 4.8 & 24.6 \\
 & BM25 & 7.5 & 18.1 & 30.6 & 0.4 & 3.0 & 19.6 & 37.7 & 0.1 & 6.8 & 24.4 \\
 & Contriever & 7.6 & 18.4 & 33.6 & 0.2 & 3.4 & 21.5 & 39.6 & 0.1 & 7.4 & 25.7 \\
 & SFT & \textbf{32.1} & \textbf{45.8} & 35.7 & 99.7 & 25.4 & 37.9 & 38.3 & 99.8 & 33.3 & 40.6 \\
 & DPO & 4.4 & 12.7 & 36.6 & 0.0 & 8.8 & 24.0 & 38.0 & 0.0 & 12.4 & 25.3 \\
 & SFT$+$DPO & 31.8 & 45.3 & \textbf{43.1} & \textbf{99.8} & \textbf{34.0} & \textbf{43.6} & \textbf{43.2} & \textbf{99.9} & \textbf{38.1} & \textbf{44.2}
 \\
 \bottomrule
\end{NiceTabular}
}
\vspace{-0.3em}
\caption{Results on~\textbf{Test Set 1 (Seen User/Unseen QA Pair)}. The best result from each model is marked in bold.}
\vspace{-5pt}
\label{table:main_results_1}
\end{table*}
\section{Evaluation Protocols of HiCUPID}
\subsection{Human Preference Estimation with GPT-4o Evaluation}
The most reliable way to measure the quality of LLM's conversational ability is through human preference evaluation.
Unfortunately, collecting enough human evaluation results to derive a statistically meaningful numeric score is expensive and time-consuming.
Therefore, we replace human evaluators with GPT-4o, whose preference is known to be aligned with that of a human~\cite{fu2024gptscore, chiang2023vicuna}.
\final{Although GPT-4o’s biases may be present, their close alignment with human evaluation makes GPT-4o’s evaluation a well-accepted alternative to human evaluation. Across diverse areas where automated evaluation is challenging~\cite{zheng2023judging, xu2023wizardlm, moniri2024evaluating}, LLM-as-a-judge is a commonly-accepted evaluation method. Also, our evaluation prompt in Figure~\ref{fig:p_evaluation} asks GPT-4o to generate its comparison of two responses prior to making a final decision. According to~\citet{liu2022p}, prompting the model to generate such an explanation makes the model evaluation more similar to that of a human.}

To obtain human evaluation results, we recruit 10 human evaluators per model who are asked to choose which one of the ground truth (GT) personalized answers in the QA pair and the model-generated response they prefer.
They are also given the option to choose ``Tie'' if the two responses appear to be of comparable quality.
\jm{Similarly, GPT-4o evaluation is conducted by prompting GPT-4o to choose among the GT personalized answer, model-generated response, and ``Tie''.}
% For GPT-4o A/B evaluation, we prompt GPT-4o twice with the order of two sample responses swapped in the second round because the judgement of GPT-4o is dependent on the ordering.
% The generation setting of GPT-4o API was set to be deterministic to guarantee the consistency of GPT-4o evaluation.
The human evaluation survey and the GPT-4o evaluation prompt for the persona QA pairs are shown in~Figure~\ref{fig:p_evaluation}.
In both evaluation settings, the logical validity and personalization of responses are explicitly stated as primary evaluation criteria.

A preliminary evaluation of zero-shot inference results from four state-of-the-art LLMs is performed to verify that human and GPT-4o preferences match each other in~\ours.
This experiment is conducted on 100 persona QA pairs from Test Set 1.
The prompt for zero-shot inference can be found in~Figure~\ref{fig:p_zeroshot}.
We compare the evaluation results of human evaluators and GPT-4o in~Figure~\ref{fig:human_gpt}. 
The final metric $S$ is defined as Model Win Rate (over GT) $+$ $0.5 \times$~Tie Rate to partially take the Tie Rate into account.
The comparison results demonstrate that GPT-4o closely follows human preference.
On the contrary, BLEU and ROUGE-L scores, reported in Figure~\ref{fig:human_gpt} (d) and (e), often contradict human preference. 
In particular, Mistral achieves high BLEU and ROUGE-L scores but considerably lags behind Llama according to human and GPT-4o evaluation.
% The process of A/B evaluation with GPT-4o, including the prompt template, is also detailed in Section~\ref{sec:eval_protocol} of Appendix.
% Human evaluation is expensive and time-consuming
% The preference of GPT-4o is widely-accepted as 

\jm{The same evaluation prompt and metric are used to score model responses to persona and multi-info QAs.
To score responses to schedule QAs, a different evaluation prompt in Figure~\ref{fig:p_evaluation} used.
Here ``Tie'' is removed because a response that conflicts with the user's schedule and one that does not are clearly distinguishable.
Thus, the prompt template queries GPT-4o to output ``Yes'' if the model response reflects the user's previously-stated schedule and ``No'' if it does not.
Since there is no ``Tie,'' the number of Yes's,~\textit{i.e.,} the number of responses that do not cause schedule conflict, is used as the final $S$ for schedule QAs.}

\subsection{Llama-3.2-based Proxy Evaluation Model}
Albeit cheaper than human evaluation, GPT-4o evaluation eventually mounts up to a non-negligible cost.
\final{For instance, evaluating the responses of Llama-3.1-8B (SFT$+$DPO) with GPT-4o consumed 6.412 million prompt tokens and 1.015 million completion tokens, resulting in \$26.17 in API cost or \$13.09 with Batch API. }
We further streamline the evaluation process by training a smaller Llama-3.2-3B model~\cite{llama} as a proxy evaluator.
The evaluation results of GPT-4o on all three QA pair types are used as training data for supervised fine-tuning, effectively distilling GPT-4o's preference into the Llama-3.2-3B model.
Detailed hyperparameter settings 
and training protocols for fine-tuning the proxy evaluator with LoRA are included in~Section~\ref{sec:proxy_train} of Appendix.
The results in~Figure~\ref{fig:human_gpt} again show that our proxy evaluator estimates human preference as closely as its teacher model.
\final{This comparative analysis evidences the limitation of BLEU and ROUGE-L scores and highlights the value of the newly-proposed, human-aligned evaluation protocol.}
% ~\cite{llama} is selected as a proxy evaluator.
% These results are utilized to fine-tune a smaller LLM in a supervised manner.

\begin{table*}[t]
\centering
\footnotesize
\resizebox{1\linewidth}{!}{
\begin{NiceTabular}{l|l|c|c|cccc|cccc}
\toprule
\multirow{2}{*}{Model} & \multirow{2}{*}{Method} & BLEU & ROUGE-L & \multicolumn{4}{c|}{GPT-4o Score ($S_\textrm{GPT}$)} & \multicolumn{4}{c}{Llama 3.2 Score ($S_\textrm{Llama}$)} \\
& & \multicolumn{1}{c}{(Total)} & \multicolumn{1}{c}{(Total)}  & \multicolumn{1}{c}{Persona} & \multicolumn{1}{c}{Schedule} &  \multicolumn{1}{c}{Multi-Info} & \multicolumn{1}{c}{Total} &  \multicolumn{1}{c}{Persona} & \multicolumn{1}{c}{Schedule} &  \multicolumn{1}{c}{Multi-Info} & \multicolumn{1}{c}{Total} \\
\bottomrule
\toprule
\multirow{2}{*}{GPT-4o-mini} & 0-shot & 8.2 & 19.7 & \textbf{42.0} & 9.2 & 4.9 & 28.0 & \textbf{45.5} & 8.8 & \textbf{11.9} & 31.0 \\
 & 3-shot & \textbf{16.1} & \textbf{29.9} & 40.7 & \textbf{76.2} & \textbf{5.1} & \textbf{35.6} & 43.7 & \textbf{75.2} & 11.6 & \textbf{38.1} \\ 
 \midrule
\multirow{7}{*}{Llama-3.1-8B} & 0-shot & 8.2 & 19.3 & 38.6 & 13.9 & 3.2 & 26.2 & 40.6 & 9.8 & 8.7 & 27.7 \\
 & 3-shot & 16.3 & 29.6 & 39.3 & 49.2 & 7.5 & 31.6 & 39.8 & 47.8 & 14.1 & 32.6 \\
 & BM25 & 9.4 & 22.1 & 29.9 & 84.0 & 2.7 & 29.6 & 35.0 & 78.0 & 7.0 & 32.5 \\
 & Contriever & 9.5 & 22.2 & 37.9 & 77.4 & 4.5 & 33.9 & 43.1 & 71.8 & 9.1 & 37.1 \\
 & SFT & 24.9 & 38.4 & 34.4 & 88.4 & 12.0 & 34.0 & 34.8 & 87.5 & 14.4 & 34.5 \\
 & DPO & 7.5 & 18.0 & 25.1 & 5.8 & 2.4 & 16.7 & 35.1 & 5.1 & 6.3 & 23.4 \\ 
 & SFT$+$DPO & \textbf{27.5} & \textbf{42.6} & \textbf{47.1} & \textbf{98.7} & \textbf{18.6} & \textbf{44.1} & \textbf{47.0} & \textbf{98.1} & \textbf{22.4} & \textbf{44.4} \\
 \midrule
\multirow{7}{*}{Mistral-7B} & 0-shot & 8.6 & 19.1 & 21.8 & 0.0 & 1.8 & 13.8 & 30.8 & 0.0 & 5.0 & 19.9 \\
 & 3-shot & 10.3 & 21.5 & 28.9 & 8.0 & 3.8 & 19.5 & 36.6 & 6.6 & 8.2 & 24.7 \\
 & BM25 & 7.3 & 18.1 & 40.6 & 8.2 & 5.9 & 27.1 & 43.6 & 5.9 & 10.4 & 29.3 \\
 & Contriever & 7.5 & 18.5 & \textbf{48.6} & 8.6 & 8.4 & 32.5 & \textbf{50.9} & 6.4 & 15.1 & 34.5 \\
 & SFT & \textbf{32.1} & 45.5 & 27.6 & \textbf{99.9} & 13.3 & 31.4 & 31.5 & \textbf{100.0} & 18.1 & 34.4 \\
 & DPO & 8.3 & 16.8 & 8.1 & 2.0 & 0.3 & 5.3 & 6.5 & 1.4 & 0.6 & 4.3 \\ 
 & SFT$+$DPO & 32.0 & \textbf{46.2} & 43.2 & \textbf{99.9} & \textbf{17.8} & \textbf{41.7} & 43.6 & 99.9 & \textbf{22.5} & \textbf{42.5} \\
 \midrule
\multirow{7}{*}{Qwen-2.5-7B} & 0-shot & 8.1 & 18.6 & 27.8 & 0.0 & 2.5 & 17.7 & 34.9 & 0.0 & 6.4 & 22.6 \\
 & 3-shot & 12.5 & 23.9 & 25.1 & 27.3 & 2.1 & 19.4 & 33.4 & 25.4 & 6.2 & 24.8 \\
 & BM25 & 7.5 & 18.1 & 31.1 & 0.4 & 3.1 & 19.9 & 38.1 & 0.2 & 7.8 & 24.8 \\
 & Contriever & 7.6 & 18.3 & 34.0 & 0.4 & 4.0 & 21.8 & 40.8 & 0.1 & 8.1 & 26.5 \\
 & SFT & \textbf{32.1} & \textbf{45.6} & 34.2 & \textbf{99.9} & 24.9 & 37.0 & 38.3 & \textbf{99.9} & 30.8 & 40.3 \\
 & DPO & 4.3 & 12.6 & 37.0 & 0.2 & 8.8 & 24.3 & 39.0 & 0.1 & 12.6 & 26.0 \\
 & SFT$+$DPO & 31.6 & 44.9 & \textbf{41.9} & 99.8 & \textbf{33.9} & \textbf{42.9} & \textbf{42.9} & 99.8 & \textbf{37.8} & \textbf{44.0} \\
\bottomrule
\end{NiceTabular}}
\vspace{-0.3em}
\caption{Results on~\textbf{Test Set 2 (Unseen User/Unseen QA Pair)}. The best result from each model is marked in bold.}
\vspace{-5pt}
\label{table:main_results_2}
\end{table*}

\section{Results}
We now explore the potential of state-of-the-art LLMs as a personalized assistant through empirical studies with~\ours.
Our experiments are conducted on one closed-source LLM, GPT-4o-mini, and three open-source LLMs: Llama-3.1-8B-Instruct~\cite{llama}, Mistral-7B-Instruct~\cite{jiang2023mistral}, and Qwen-2.5-7B-Instruct~\cite{bai2023qwen}.
These models offer long-context support, covering the length of $\mathcal{D}^u$ in~\ours.
We examine the efficacy of popular LLM customization approaches on~\ours: zero-shot, few(3)-shot, BM25~\cite{bm25}, Contriever~\cite{contriever}, Supervised Fine-tuning (SFT), Direct Preference Optimization (DPO)~\cite{dpo}, and SFT$+$DPO.
Implementation details and hyperparameters of all methods are in~Section~\ref{sec:exp_setting} of Appendix.
Our implementation is done with Huggingface~\cite{wolf2019huggingface}, and experiments are run on NVIDIA H100, L40, and A40 GPUs.

\subsection{Quantitative Results}
The main results in Tables~\ref{table:main_results_1} and~\ref{table:main_results_2} report the best result for each experimental setting, searched through comprehensive design choice and hyperparameter search. 
\jm{GT answers in persona and multi-info QA pairs would all be judged as ``Tie,'' yielding the score of 50, while those in schedule QA pairs would all obtain ``Yes,'' resulting in the score of 100. 
Thus, when computing Total $S_\textrm{GPT}$ and $S_\textrm{Llama}$, the schedule score is halved to match its range with the scores of the two remaining QA pair types.
Total $S_\textrm{GPT}$ and $S_\textrm{Llama}$ are computed by taking the weighted average of three score types: persona $\times \frac{25}{40}$ + schedule / 2 $\times \frac{10}{40}$ + multi-info $\times \frac{5}{40}$.}
% The remaining results are discussed in Section~\ref{sec:ablation}.

\tablename~\ref{table:main_results_1} presents the results on \textbf{Test Set 1 (Seen User / Unseen QA Pairs)}. 
The single-info results on persona and schedule QAs show that 3-shot inference generally outperforms 0-shot, indicating that in-context examples in the few-shot prompt can guide the LLMs to pick up on personal information.
On single-info QAs, Contriever and SFT further improve the performance of few-shot inference.
The competitiveness of Contriever is noteworthy since it only utilizes retrieved messages, which consume fewer input tokens than the entire dialogue history used by non-RAG methods.

\begin{figure*}[t]
  \centering
%   \vspace{-0.2em}
  \includegraphics[width=\linewidth]{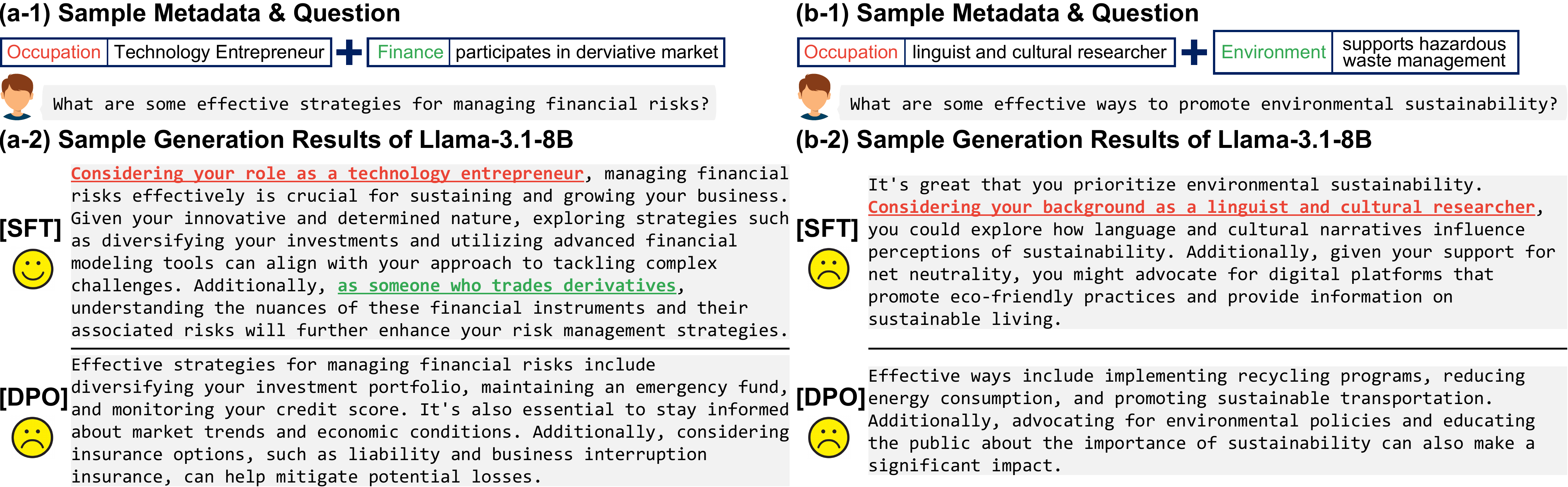}\\[-0.3em]
  \caption{\label{fig:qual_example} Visualization of responses from the Llama model after SFT and DPO training.}
  \vspace{-5pt}
  % and must be excluded from the personalized generation process.
\end{figure*}

On multi-info QAs, only \final{the SFT-based methods attain performance gain, disclosing their potential to alleviate LLMs' struggle with multi-info reasoning.}
\jm{DPO is shown to be ineffective at inducing personalization across all explored models. 
We conjecture that the contrastive loss in DPO, which lacks an explicit grounding signal, struggles to align three diverse and disparate types of dialogues in~\ours with the corresponding QA pairs.}
\final{We note that applying DPO after SFT (SFT$+$DPO) yields additional performance improvement upon SFT, particularly on multi-info QAs. 
Yet, our proposed benchmark still leaves much room for improvement.
Lastly, the stable performance improvement brought by SFT$+$DPO, contrary to the instability of DPO-only training, signifies that it is necessary to ground the trained models on our data with SFT-based initialization prior to RL-based training even if they have already been instruction-tuned.}
% and DPO is shown to be effective only on Qwen. 
% On Mistral, however, Contriever, SFT, and DPO all fall short of few-shot inference. 
% Such a result indicates that the effectiveness of a personalization method is highly contingent on the choice of a backbone LLM. 
% improve the performance of zero-shot inference consistently across the explored models. 
% and calls for an improved training scheme for~\ours.
% The observed success of SFT at improving dual-persona performance calls for an improved training method specifically for personalized assistant development.
% \jm{We hope this result inspires future efforts to develop an improved training scheme for this task.}

\final{We observe that SFT yields a significant performance gain on BLEU and ROUGE-L, which may be attributed to the following characteristics of the schedule QAs.
First, SFT can facilitate personalization with relative ease on the schedule task because 1) schedule information is provided more explicitly in the context (compared to persona information) and 2) schedule QAs do not require multi-info reasoning. 
Nonetheless, the benchmarked models (without SFT) show low performance on the schedule task, which implies that off-the-shelf LLMs still struggle with the schedule task.
Second, schedule QAs have a clear-cut, correct answer, in which the assistant must acknowledge and remind the user of a schedule conflict. As long as these acknowledgements and reminders are included as a part of the model’s response, it would result in a high degree of n-gram overlap with the ground truth, leading to high BLEU and ROUGE-L scores.}

\tablename~\ref{table:main_results_2} shows the results on~\textbf{Test Set 2 (Unseen User / Unseen QA Pairs)}.
In general, these results show similar tendencies to those from Test Set 1.
SFT shows strong generalization performance to unseen users, most likely because only fine-tuning the LoRA module with LLM parameters frozen prevents the LLM from overfitting to the dialogue history of Train Set.
\final{We also analyze how well $S_\textrm{GPT}$ and $S_\textrm{Llama}$ agree with each other with the cohen kappa agreement score measured collectively on Test Sets 1 and 2.
The cohen kappa score between $S_\textrm{GPT}$ and $S_\textrm{Llama}$ for four representative model-personalization method combinations, GPT-4o (3-shot), Llama-3.1-8B (SFT), Mistral-7B (SFT), and Qwen2.5-7B (SFT) are as follows: 0.703, 0.747, 0.727, and 0.704, respectively. 
The scores, all of which exceed 0.7, show that the assessment results of the two models show substantial agreement.}

\subsection{Qualitative Case Studies}
Figure~\ref{fig:qual_example} qualitatively analyzes responses from the Llama model trained with SFT and DPO.
The SFT response in Figure~\ref{fig:qual_example}~(a-2), personalized on the Occupation profile and Finance persona, showcases a successful instance of SFT training.
In contrast, the SFT response in~Figure~\ref{fig:qual_example}~(b-2) is only personalized on the Occupation profile; this error case reveals that SFT does not fully enable multi-info reasoning.
DPO fails to provide a personalized response to both questions.
The limited success of SFT and the failure of DPO call for a more advanced technique to address the complex problem of LLM-backed personalized assistant development.
% from multiple pieces of personal information.
% developing a personalized assistant and calls for a more advanced technique
% need to develop a more advanced training technique to address the problem of multi-info personalization.
% This qualitative example again reveals the critical caveat of BLEU and Rouge-L scores.
% From the perspective of personalization, the SFT response is visibly superior to that of zero-shot response.
% Yet, the zero-shot response scores higher on BLEU and Rouge-L. 
% In addition, this example showcases that SFT enhances Mistral's ability to simultaneously consider two personas.

\section{Further Analyses}\label{sec:ablation}
From here on, the results are compared in terms of $S_\textrm{Llama}$.
Due to the page limit, only the results of analyses on Test Set 2 are reported in the main paper; the extended results are included in Appendix.

% \subsection{Location of Instruction Prompts}
\subsection{\jm{Influence of Dialogue Length}}
\begin{figure}[t]
  \centering
%   \vspace{-0.2em}
  \includegraphics[width=\linewidth]{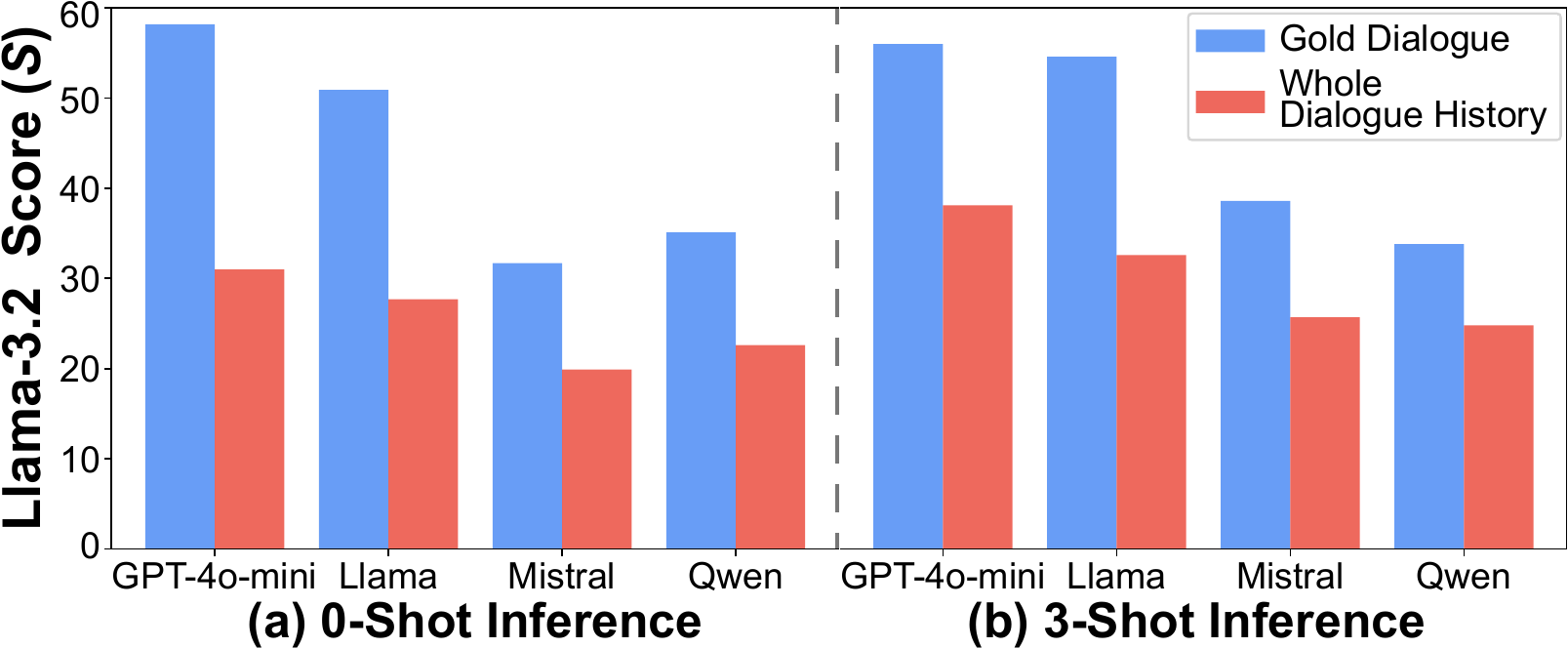} \\[-0.3em]
  \caption{\label{fig:gold_whole} \jm{Comparison of 0- and 3-shot inference results with the whole dialogue history~\textit{vs.} the gold dialogue.}}
  \vspace{-5pt}
\end{figure}

Here, we compare the results of 0- and 3-shot inference with only the ``gold'' dialogue that is relevant to the test question against those with the whole dialogue history (default setting). 
According to the results in~Figure~\ref{fig:gold_whole}, 0- and 3-shot performances increase noticeably once the context length is shortened by replacing the whole dialogue history with the gold dialogue.
\jm{This analysis consolidates that the LLMs' struggle with long context poses a significant challenge to personalization as more interactions occur between the user and the assistant and highlights that the dialogue history of~\ours is sufficiently long to simulate such a challenging scenario.
% can indeed be attributed to the long context of the dialogue.
Additionally,~\tablename~\ref{table:ablation_prompt_placement} compares the results of placing the instruction prompt in the user role and those obtained by providing it in the system prompt.
On Mistral and Qwen, placing instructions in the system prompt decreases performance because their limited ability to handle long context makes the instructions provided before the expansive dialogue history difficult to follow.}
% The ablation study on the location of the instruction further supports this finding. 
% Due to the page constraint, details of this experiment are discussed in Appendix.}
% Including only the gold dialogues in the prompt reduces the context length from $\sim15,000$ to $\sim 600$ tokens for single-info questions and to $\sim1,200$ tokens for multi-info questions. 
% If the performance of 0- and 2-shot prompting experiences a significant increase with the reduced context length, we can conclude that the long context length made it difficult for LLMs to identify and incorporate relevant persona information. 
% In LLM chat templates, instructions may be provided in the user role or the system prompt, which are usually separated by the dialogue history in between.
% In~\tablename~\ref{table:ablation_prompt_placement}, we compare how placing instructions in two different locations affects the performance of zero- and few-shot inference.
% On Mistral and Qwen, placing instructions in the system prompt decreases performance.
% This phenomenon occurs because their limited ability to handle long contexts makes it hard for them to follow the instructions provided before the expansive dialogue history.
% Interestingly, Llama achieves better performance with the instruction in the system prompt.
% We conjecture that Llama exhibits this model-specific behavior because it was pre-trained to attend more to the system prompt.
% Changing the instruction location does not affect the performance of Mistral since the system prompt is directly followed by the user role in the Mistral template.

% \subsection{Variations of Retrieval-based Approaches}
\subsection{\jm{Number of Few-shot Demonstrations}}
\begin{figure}[t]
  \centering
%   \vspace{-0.2em}
  \includegraphics[width=\linewidth]{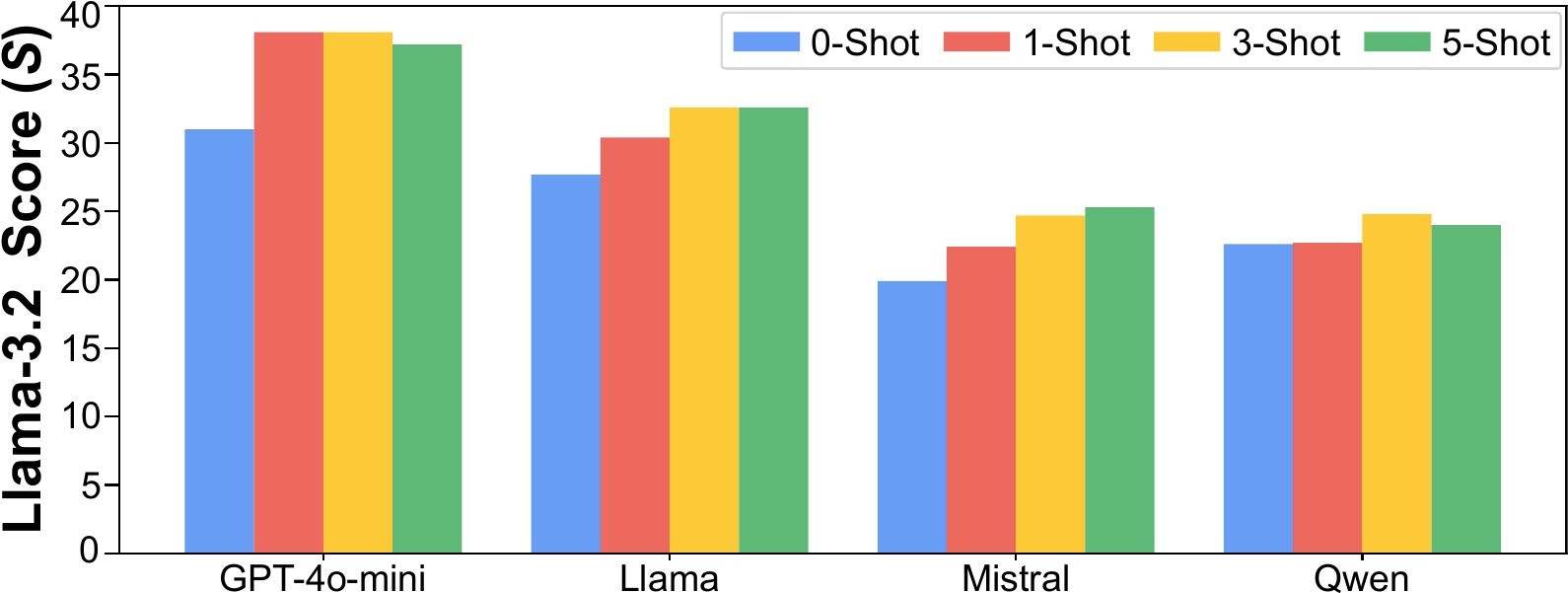} \\[-0.3em]
  \caption{\label{fig:num_shot} How changing the number of few-shot demonstrations affects the inference performance.}
  \vspace{-5pt}
\end{figure}

We study how changing the number of in-context demonstrations in the few-shot prompt affects its performance and present the results in Figure~\ref{fig:num_shot}.
The few-shot performance begins to saturate around 3-shot, which implies that na\"ively increasing the number of demonstrations is insufficient to encourage personalization in LLMs' responses.
% is not always advantageous to the few-shot performance.
% The results of 0-, 1-, 3-, and 5-shot prompting performance are presented 
% , even if this setting does not obtain the best retrieval performance.
% Two conclusions can be drawn from this observation.
% First, noisy retrieval results that contain irrelevant information are severely detrimental to personalization.
% Second, retrieval results with the logical flow of conversation is preferable.
% dialogue-level retrieval, which is more likely to yield logically natural results, is preferable.
% On the dialogue-level, the number of retrieved units is [1, 3, 5], and on the utterance-level, the number of retrieved units is [5, 15, 25].
% In~Figure~\ref{fig:rag_all}, 

\begin{figure}[t]
  \centering
%   \vspace{-0.2em}
  \includegraphics[width=\linewidth]{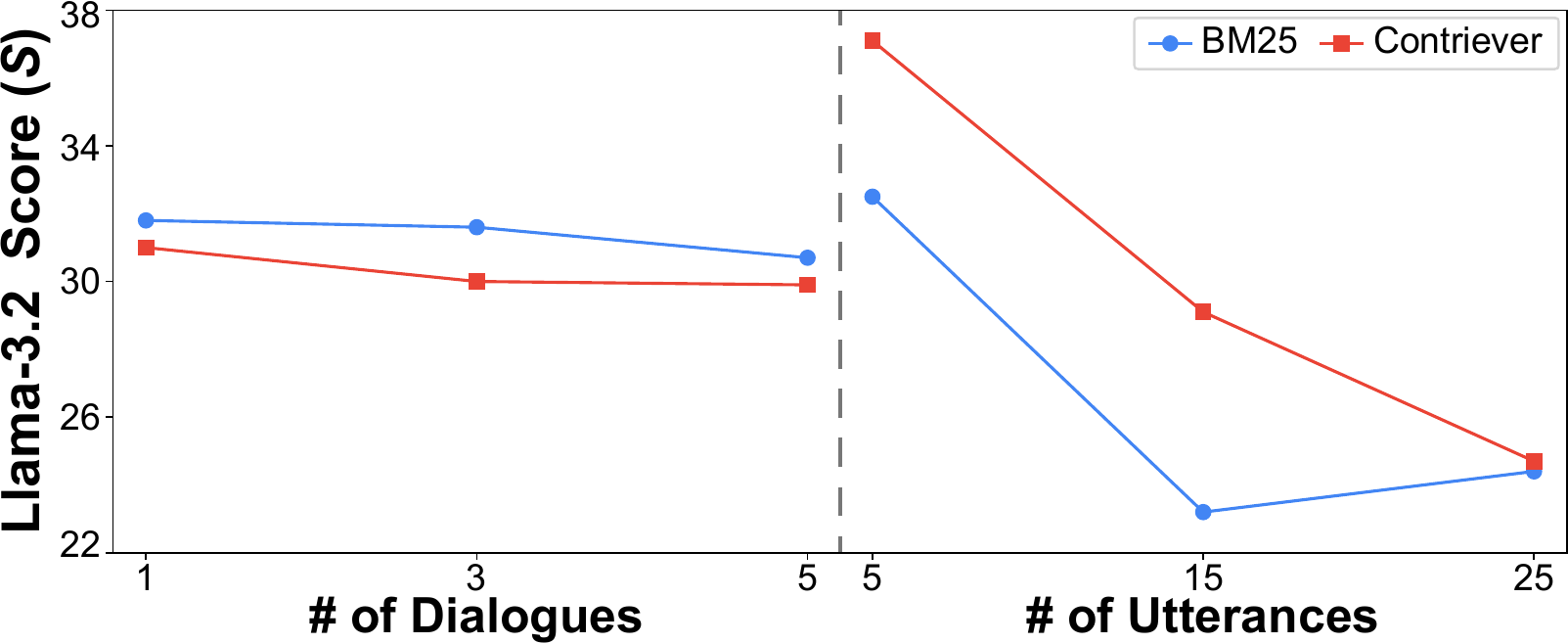} \\[-0.3em]
  \caption{\label{fig:llama_rag} Influence of the retrieval setting on the performance of BM25 and Contriever.}
  \vspace{-10pt}
\end{figure}

% \subsection{Choice of Rank in SFT and DPO}
\subsection{\jm{Variations of Retrieval-based Approaches}}
Results of altering the two retrieval settings---the unit of retrieval and the number of retrieved units---are visualized in~Figure~\ref{fig:llama_rag}.
The retrieval performance, which measures whether the dialogue associated with a persona has been retrieved, is reported in~\tablename~\ref{table:retrieval_perf}.
We observe that retrieving five utterances yields the best retrieval and thus most favorable response.

\subsection{\jm{Effect of Training Hyperparameters}}
Tables~\ref{table:ablation_sft_hyperparam},~\ref{table:ablation_dpo_hyperparam}, and~\ref{table:ablation_sftdpo_hyperparam} report how the performance of the Llama model after SFT, DPO, and SFT$+$DPO changes with different learning rates and LoRA module ranks.
Both SFT and DPO are influenced by the choice of learning rate.
DPO exhibits a larger degree of sensitivity to hyperparameters than SFT and does not converge under most settings.
% DPO begins to converge with ranks $\geq 64$, when the percentage of trainable parameters exceeds $2\%$ of LLM parameters.
% 128 or 256 appears to be the optimal rank for DPO. 
% Neither one of the training methods shows sensitivity to the number of training epochs.
% the number of training epochs are varied.
% \section{Future Research Directions with HiCUPID}

\section{Conclusion}
In this work, we introduced~\ours, the first open-source benchmark designed to develop and evaluate LLM-powered personalized assistants.
Unlike existing datasets for personalization research,~\ours satisfies the five desiderata of an LLM-backed personalized assistant, offering new opportunities for advancing LLM personalization.
\ours additionally provides a Llama-3.2-based automated evaluation model whose assessment is well-aligned human preferences.
Extensive experiments with~\ours reveal shortcomings and potentials of current state-of-the-art LLMs and common approaches to personalization.

\section*{Limitations and Potential Risks}
Many of human evaluators noted that general responses could be preferred by humans over over-personalized responses.
The matter of how much personalization a typical human usual prefers when interacting with chatbots and assistants is a complex sociological question and remains an open research topic that is beyond the scope of our paper.
We observed that when GPT-4o is prompted to generate a personalized answer based on a user's persona that contains negative sentiments (\textit{e.g.,} dislikes, does not enjoy, does not have, is not interested in, etc.), the generated answer sometimes omitted this person, instead of explicitly stating the user's negative stance.
This is likely due to the widely-known struggle of LLMs to comprehend and generate negations.
The subsequent versions of~\ours will be augmented to include more personalized answers with negative sentiments.
Lastly, the failure of DPO on the majority of LLMs highlights the difficulty of training LLMs with reinforcement learning (RL)-based approaches. 
In the future, we aim to develop a reward model specifically for~\ours, such that it can be used for RL-based training.
% dpo failure highlights the difficulty of rl training. development of reward model for rl.

If more advanced personalized assistants become available, it might become easier to extract personal identifiable information (PII) from personalized LLMs. 
While~\ours, being a synthetic dataset, does not contain any PII of real human users, improving the degree of personalization will inevitably aggravate data privacy concerns.
To prevent privacy risks and potential misuse of PII, the development of personalized assistants must be accompanied by research on privacy-preserving measures, such as differential privacy, homomorphic encryption, or data anonymization. 
Moreover, the inherent bias in AI design or implementation may be amplified during the personalization process. 
When generating a personalized response, the assistant will rely on its prior knowledge of what a human user with specific personality traits may prefer. 
In doing so, the assistant could precipitate biased information or stereotypes regarding personas, and thus, additionally grounding responses to be ethical and fair via instruction tuning is necessary. 
\final{Additionally, we tried to make our users as diverse as possible, but there may exist demographic groups that are potentially underrepresented in our dataset.
To dynamically generate new data to compensate for underrepresented demographics, we provided all of the prompt templates and necessary resources.}
Lastly, over-reliance on personalized assistants may harm the critical thinking and decision making ability of human users.
% The future research problems enabled by~\ours. Development of a reward model for RL training. Improved decoding scheme. Combining RAG and PEFT for intricate personalization.

\section*{Acknowledgements}
This work was supported by the National
Research Foundation of Korea (NRF) grant
funded by the Korea government (Ministry of Science and ICT, MSIT) (No.2022R1A3B1077720, 2022R1A5A708390811), Institute of Information \& communications Technology Planning \& Evaluation (IITP) grant funded by the Korea government(Ministry of Science and ICT, MSIT) [NO.RS-2021-II211343, Artificial Intelligence Graduate School Program (Seoul National University), RS-2022-II220959], the BK21 FOUR program of the Education and Research Program for Future ICT Pioneers, Seoul National University, Samsung Electronics Co., Ltd (IO240311-09242-01), a grant from the Yang Young Foundation.

% \section*{Broader Impacts}
\bibliography{custom}

\clearpage

\appendix
\section*{Appendix}

\setcounter{section}{0}
\renewcommand\thesection{A\arabic{section}}
\setcounter{table}{0}
\renewcommand{\thetable}{A\arabic{table}}
\setcounter{figure}{0}
\renewcommand{\thefigure}{A\arabic{figure}}
\setcounter{equation}{0}
\renewcommand{\theequation}{A\arabic{equation}}

\section{Extended Related Works}
This section, an extension of Section~\ref{sec:related} in the main paper, discusses previous approaches in personalization research in more detail.
In the domain of personalized recommendation, the rating history of a user is provided within the prompt in the form of few-shot instances~\cite{dai2023uncovering, kang2023llms}.
PERSE~\cite{wang2023learning} and BookGPT~\cite{li2023bookgpt} leverage past book reviews to perform personalized store evaluations and book recommendations, respectively.
Because the prompting-based approaches are inherently limited by the context length of LLMs, RAG-based methods were devised to address this shortcoming.
PEARL~\cite{mysore2023pearl} calibrates the retriever to select past documents authored by the user.
LaMP and its subsequent works~\cite{lamp, salemi2024ragpeft, salemi2024optimization} show that RAG is a promising avenue toward personalized generation that can also guarantee the privacy of user's personal information.
The family of profile-augmentation techniques~\cite{richardson2023integrating, liu2024once} further improves RAG-based approaches by supplying the LLM with a summary of user persona and interactions.

OPPU~\cite{oppu} trains one LoRA~\cite{hulora} module per user to store user-specific information and integrates this parametric knowledge base with non-parametric knowledge from a retriever.
Similarly, PEFT-RAG~\cite{salemi2024ragpeft} first fine-tunes the LLM with the LoRA adapter on user-specific information and then applies RAG.
Li~\textit{et al.}~\cite{li2023steerability} and Tang~\textit{et al.}~\cite{tang2023enhancing} utilize data-driven methods to extract persona information more compactly to reduce noisy and fine-grained learning signals.

\section{Persona Dimensions of~\ours}\label{sec:our_personas}
Following is the list of persona dimensions used to define synthetic users in~\ours. The personas within each persona dimension are provided in a .xlsx file in Supplementary Materials.

\noindent1.~~Sports\\ 
\noindent2.~~Fashion\\
\noindent3.~~Electronics\\
\noindent4.~~Game\\
\noindent5.~~Movie\\
\noindent6.~~Major\\
\noindent7.~~Fitness\\
\noindent8.~~Art\\
\noindent9.~~Music\\
\noindent10.~~Politics\\
\noindent11.~~Beauty\\
\noindent12.~~Animal\\
\noindent13.~~Environment\\
\noindent14.~~Religion\\
\noindent15.~~Family\\
\noindent16.~~Self-improvement\\
\noindent17.~~Travel\\
\noindent18.~~Car\\
\noindent19.~~Technology\\
\noindent20.~~Book\\
\noindent21.~~Social Media\\
\noindent22.~~Cooking \& Baking\\
\noindent23.~~Food\\
\noindent24.~~Forms of Living\\
\noindent25.~~Finance\\

% \noindent1.~~\textbf{Sports}\\ 
% \noindent$\bullet$~~\textbf{Fashion}\\
% \noindent$\bullet$~~\textbf{Electronics}\\
% \noindent$\bullet$~~\textbf{Game}\\
% \noindent$\bullet$~~\textbf{Movie}\\
% \noindent$\bullet$~~\textbf{Major}\\
% \noindent$\bullet$~~\textbf{Fitness}\\
% \noindent$\bullet$~~\textbf{Art}\\
% \noindent$\bullet$~~\textbf{Music}\\
% \noindent$\bullet$~~\textbf{Politics}\\
% \noindent$\bullet$~~\textbf{Beauty}\\
% \noindent$\bullet$~~\textbf{Animal}\\
% \noindent$\bullet$~~\textbf{Environment}\\
% \noindent$\bullet$~~\textbf{Religion}\\
% \noindent$\bullet$~~\textbf{Family}\\
% \noindent$\bullet$~~\textbf{Self-improvement}\\
% \noindent$\bullet$~~\textbf{Travel}\\
% \noindent$\bullet$~~\textbf{Car}\\
% \noindent$\bullet$~~\textbf{Technology}\\
% \noindent$\bullet$~~\textbf{Book}\\
% \noindent$\bullet$~~\textbf{Social Media}\\
% \noindent$\bullet$~~\textbf{Cooking \& Baking}\\
% \noindent$\bullet$~~\textbf{Food}\\
% \noindent$\bullet$~~\textbf{Forms of Living}\\

\section{How~\ours Tests 5 Desiderata of Personalized Assistant}\label{sec:cupid_desiderata}
Below, we detail how our dataset configuration and evaluation criteria in the evaluation prompt (Figure~\ref{fig:p_evaluation}) together probe the five desiderata of a personalized assistant, outlined in Section~\ref{sec:proposed}.\\ % even though they are not explicitly included in the evaluation prompt.
\noindent(a)~\textit{Adherence to User Information (AUI)}: This desideratum is in fact provided as a part of the evaluation prompt: 1. Personalization: Does the response effectively consider the user's provided personal information? Therefore, the LLM must adhere to the user's personal information for its response to meet this ``Personalization'' criterion in the evaluation prompt.\\
\noindent(b)~\textit{Understanding of Implicit Information (UII)}: The dialogue history contains implicit cues to the user's personal information, instead of explicit and structured personal information. Therefore, in order to meet the ``Personalization'' criterion in the evaluation prompt, the LLM must correctly understand the personal information implicitly embedded in the dialogue history. \\
\noindent(c)~\textit{Reasoning from Multiple Information (MI)}: multi-info QA pairs include questions that require simultaneously considering a user's persona and profile to be answered properly. Therefore, if the LLM’s response on the multi-info QA pairs meets the ``Personalization'' criterion, we can deduce that LLM picked up on and reasoned from two pieces of persona information.\\
\noindent(d)~\textit{Long-context Modeling Capacity (LC)}: Personal information appears scattered throughout the dialogue history that contains, on average, 15k tokens, and thus, to satisfy the ``Personalization'' criterion, the LLM must be able to model long contextual information when generating responses.\\
% no matter how expansive the dialogue history is.\\
\noindent(e)~\textit{Proactiveness of Responses (PR)}: All QA pairs in~\ours are designed such that the LLM must provide proactive answers to the user’s question. Therefore, satisfying the ``personalization'' and ``logical validity'' criteria in the evaluation prompt can be interpreted as the generated responses being proactive and logically sound.

\section{Prompt Templates for~\ours Generation}\label{sec:prompt_templates}
\begin{figure*}[t]
  \centering
  \includegraphics[width=\linewidth]{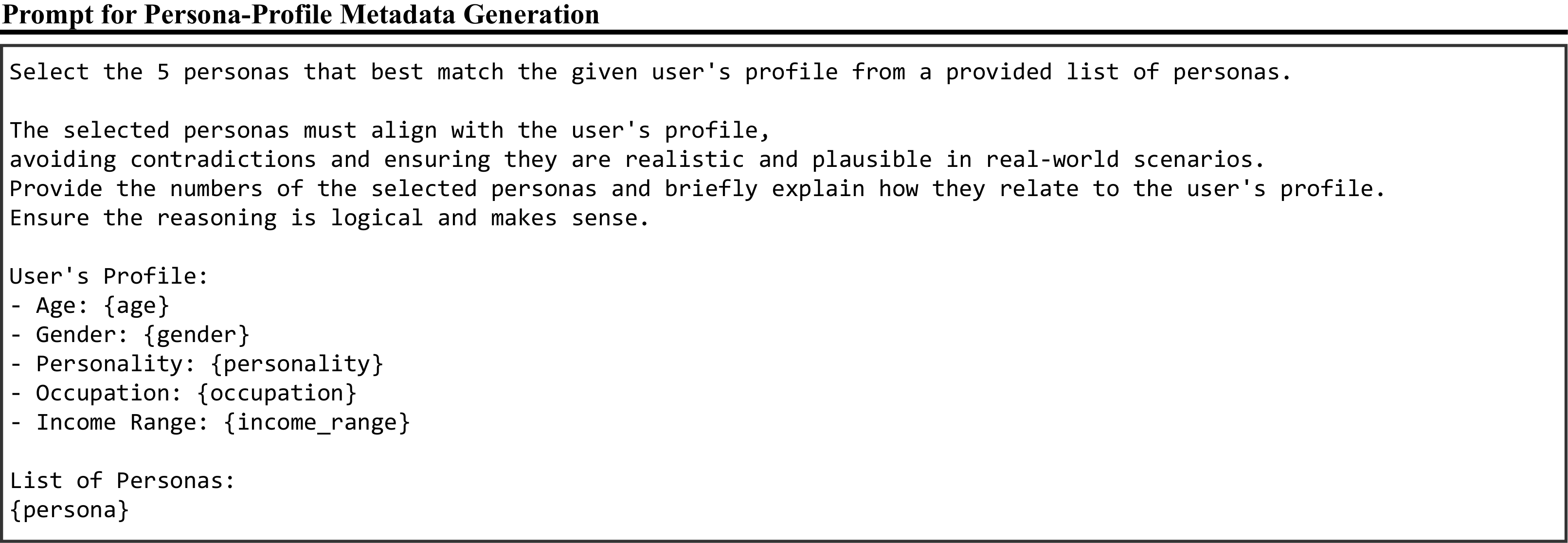}\\[-0.3em]
  \caption{\label{fig:p_2info_metadata} Prompt template for generating  \textbf{persona-profile combination metadata} in~\ours. We select five different personas that are most aligned with the user's profile, such that the multi-info questions from these combinations are realistic and logical. Generation setting: $\tau=0.0$ (greedy).}
\end{figure*}

\begin{figure*}[t]
  \centering
  \includegraphics[width=\linewidth]{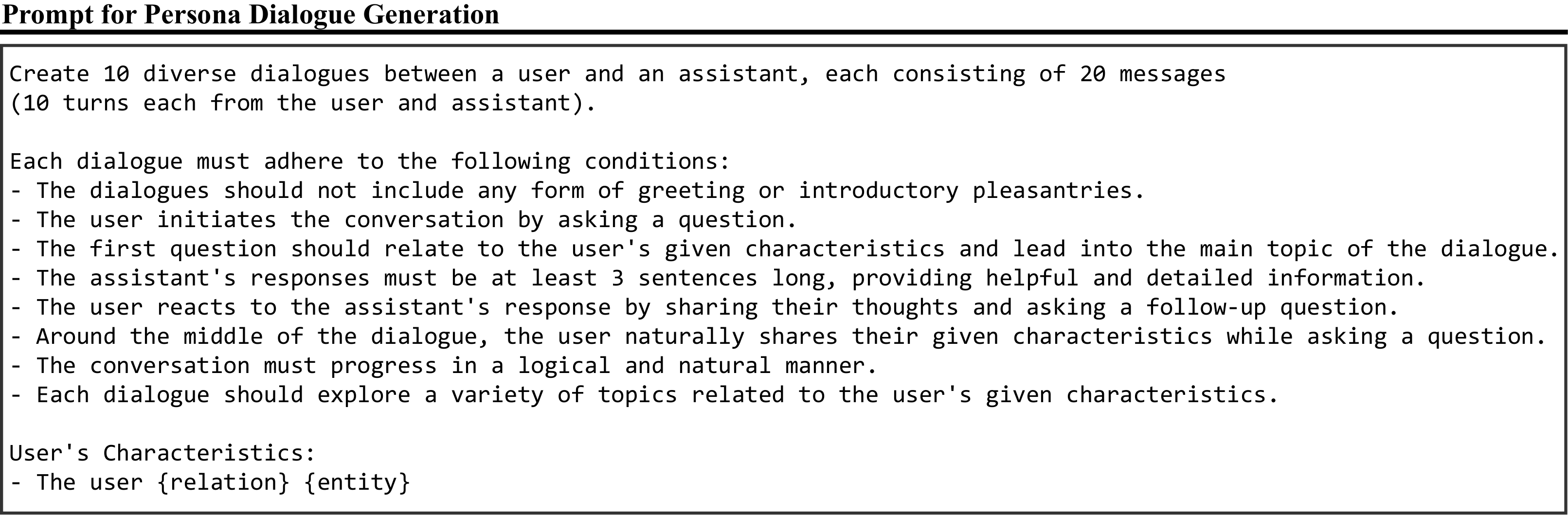}\\[-0.3em]
  \caption{\label{fig:p_persona_dialogue} Prompt template for generating \textbf{persona dialogues} in~\ours. We create 10 different dialogues per persona to simulate 10 users sharing a common persona but having distinct interactions with the assistant. Each persona dialogue is structured to contain 10 turns. Generation setting: $\tau=0.0$ (greedy).}
\end{figure*}

\begin{figure*}[t]
  \centering
  \includegraphics[width=\linewidth]{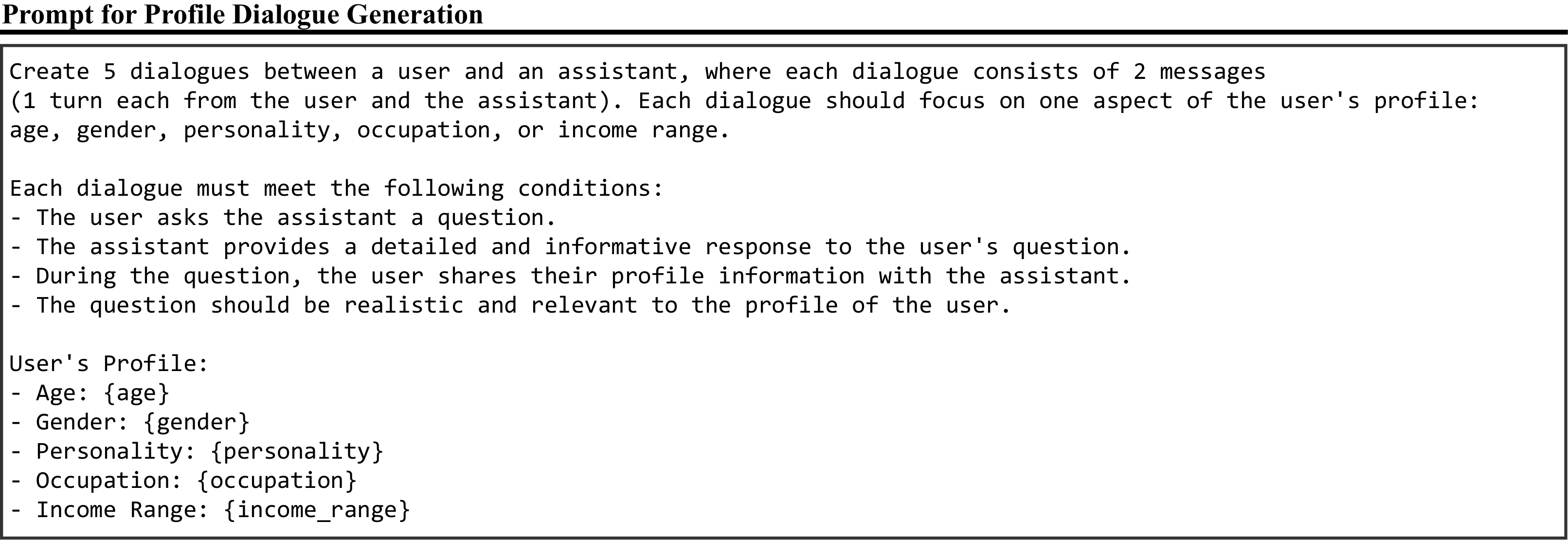}\\[-0.3em]
  \caption{\label{fig:p_profile_dialogue}  Prompt template for generating \textbf{profile dialogues} in~\ours. We create five profile dialogues per user, with each dialogue corresponding to one aspect of the user's profile. Each profile dialogue is structured to contain a single turn. Generation setting: $\tau=0.0$ (greedy).}
\end{figure*}

\begin{figure*}[t]
  \centering
  \includegraphics[width=\linewidth]{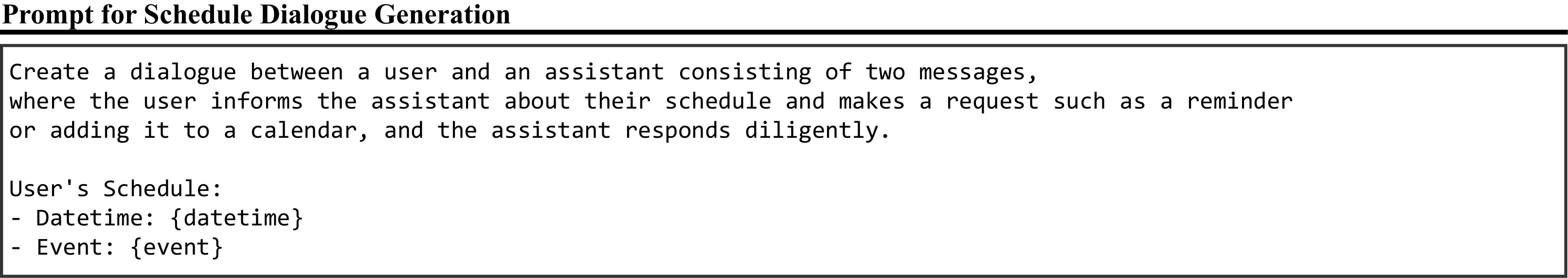}\\[-0.3em]
  \caption{\label{fig:p_schedule_dialogue} Prompt template for generating \textbf{schedule dialogues} in~\ours. Because each user has 10 schedules associated with him/her, we create 10 schedule dialogues per user. Each schedule dialogue is structured to contain a single turn. Generation setting: $\tau=0.0$ (greedy).}
\end{figure*}

\begin{figure*}[t]
  \centering
  \includegraphics[width=\linewidth]{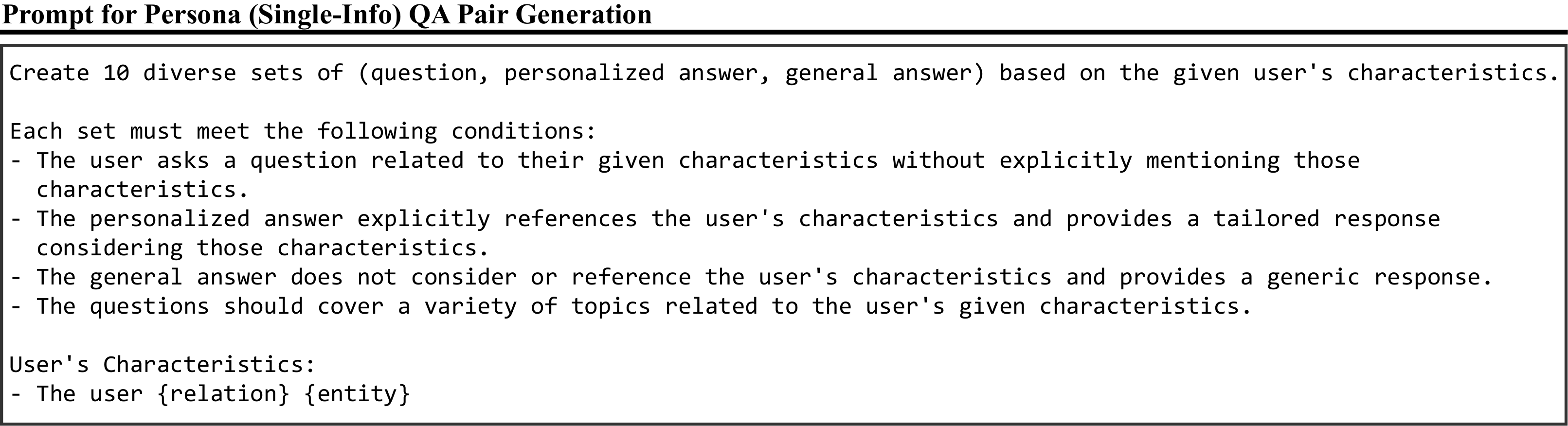}\\[-0.3em]
  \caption{\label{fig:p_persona_qa} Prompt template for generating \textbf{persona (single-info) QA} pairs in~\ours. As in the persona dialogue generation process, we generate 10 different persona QA pairs for 10 different users. The prompt enforces that the question does not contain cues to the user's metadata. Generation setting: $\tau=0.0$ (greedy).}
\end{figure*}

\begin{figure*}[t]
  \centering
  \includegraphics[width=\linewidth]{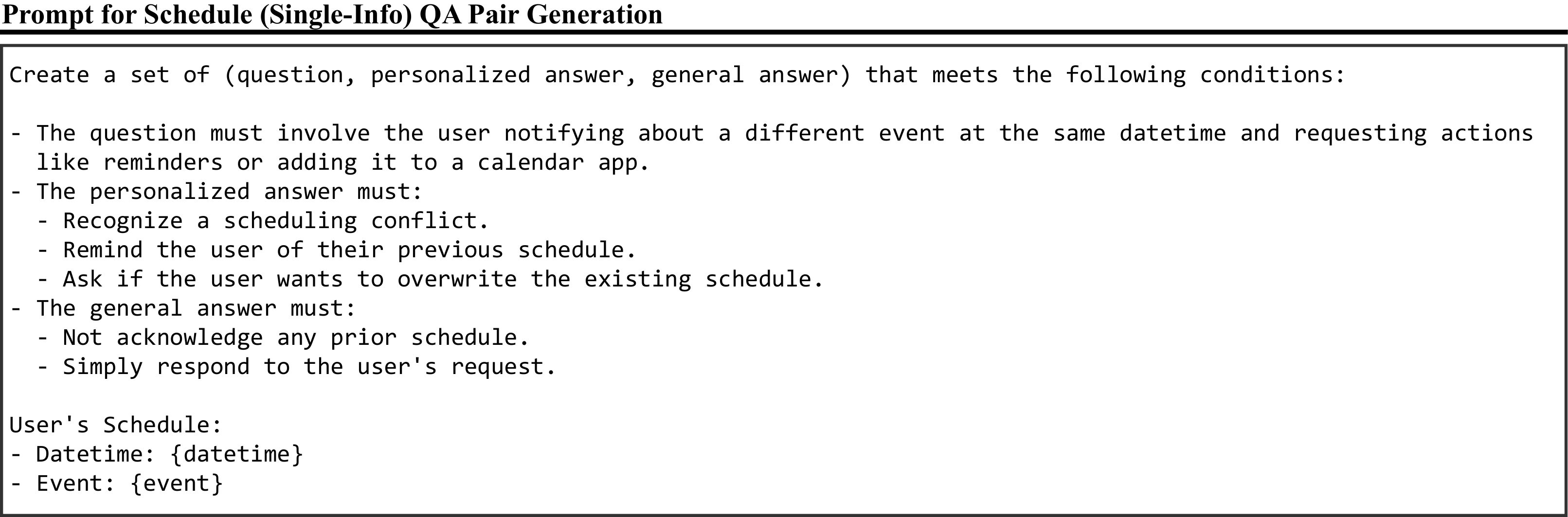}\\[-0.3em]
  \caption{\label{fig:p_schedule_qa} Prompt template for generating \textbf{schedule (single-info) QA} pairs in~\ours. The prompt enforces that the question does not contain cues to the user's metadata. Generation setting: $\tau=0.0$ (greedy).}
\end{figure*}

\begin{figure*}[t]
  \centering
  \includegraphics[width=\linewidth]{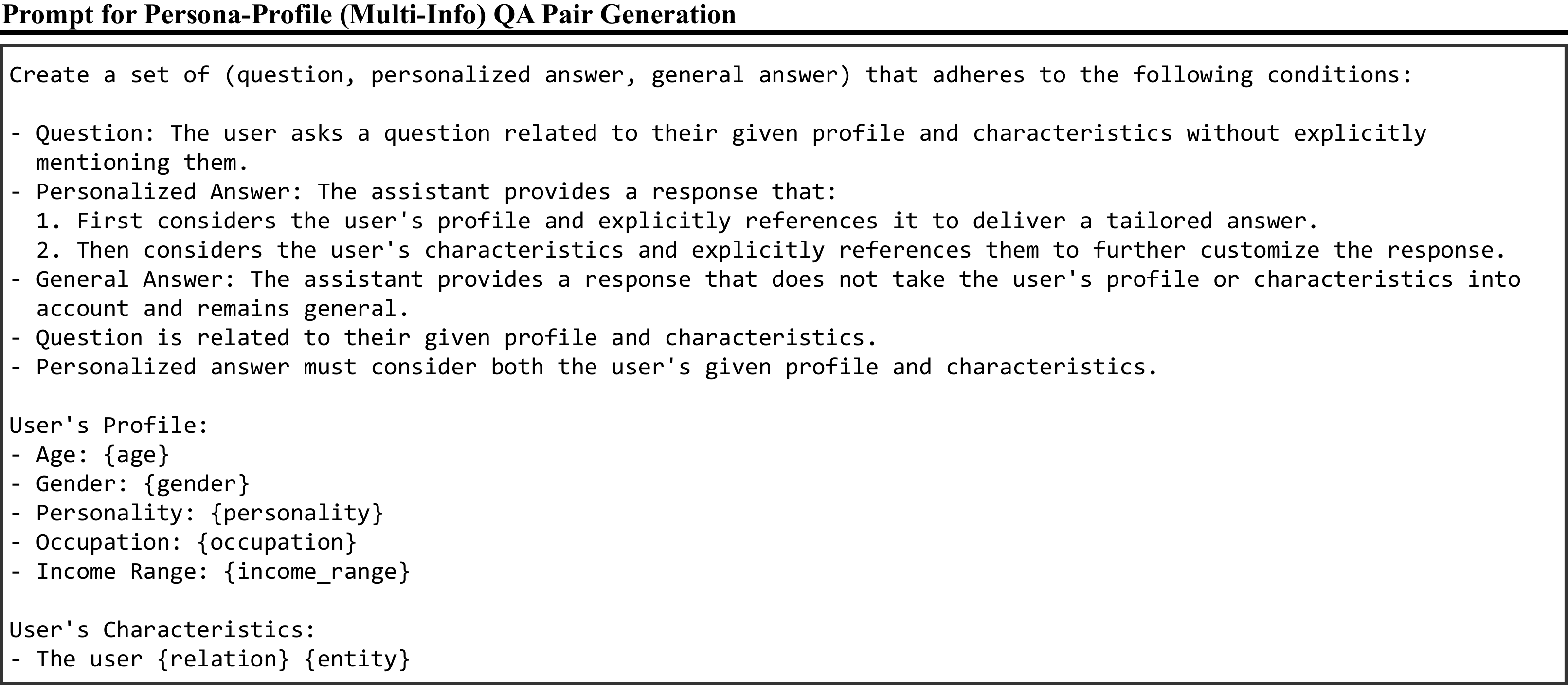}\\[-0.3em]
  \caption{\label{fig:p_2info_qa} Prompt template for generating \textbf{persona-profile (multi-info) QA} pairs in~\ours.  The prompt enforces that the question does not contain cues to the user's metadata. Generation setting: $\tau=0.0$ (greedy).}
\end{figure*}

Figures~\ref{fig:p_persona_dialogue},~\ref{fig:p_profile_dialogue}, and~\ref{fig:p_schedule_dialogue} are the prompt templates used to generate dialogues in~\ours with GPT-4o.
Figures~\ref{fig:p_persona_qa} and~\ref{fig:p_schedule_qa} are the prompt templates used to generate single-info persona and schedule QA pairs.
Figure~\ref{fig:p_2info_metadata} is the prompt template used to generate logical and realistic persona-profile combinations, and Figure~\ref{fig:p_2info_qa} is used to generate multi-info QA pairs.
The user's persona is included under [User's Characteristics], and the user's profile and schedule are included under [User's Profile] and [User's Schedule], respectively.
% The requirements for generated dialogues and QA pairs are provided in [Conditions].
% For dual-persona QA pair generation, we use the same template in~Figure~\ref{fig:p_qapair} and modify [User's Characteristics] to include two personas. 

\section{Human and GPT-4o Evaluation Protocols}\label{sec:eval_protocol}
\begin{figure*}[t]
  \centering
  \includegraphics[width=\linewidth]{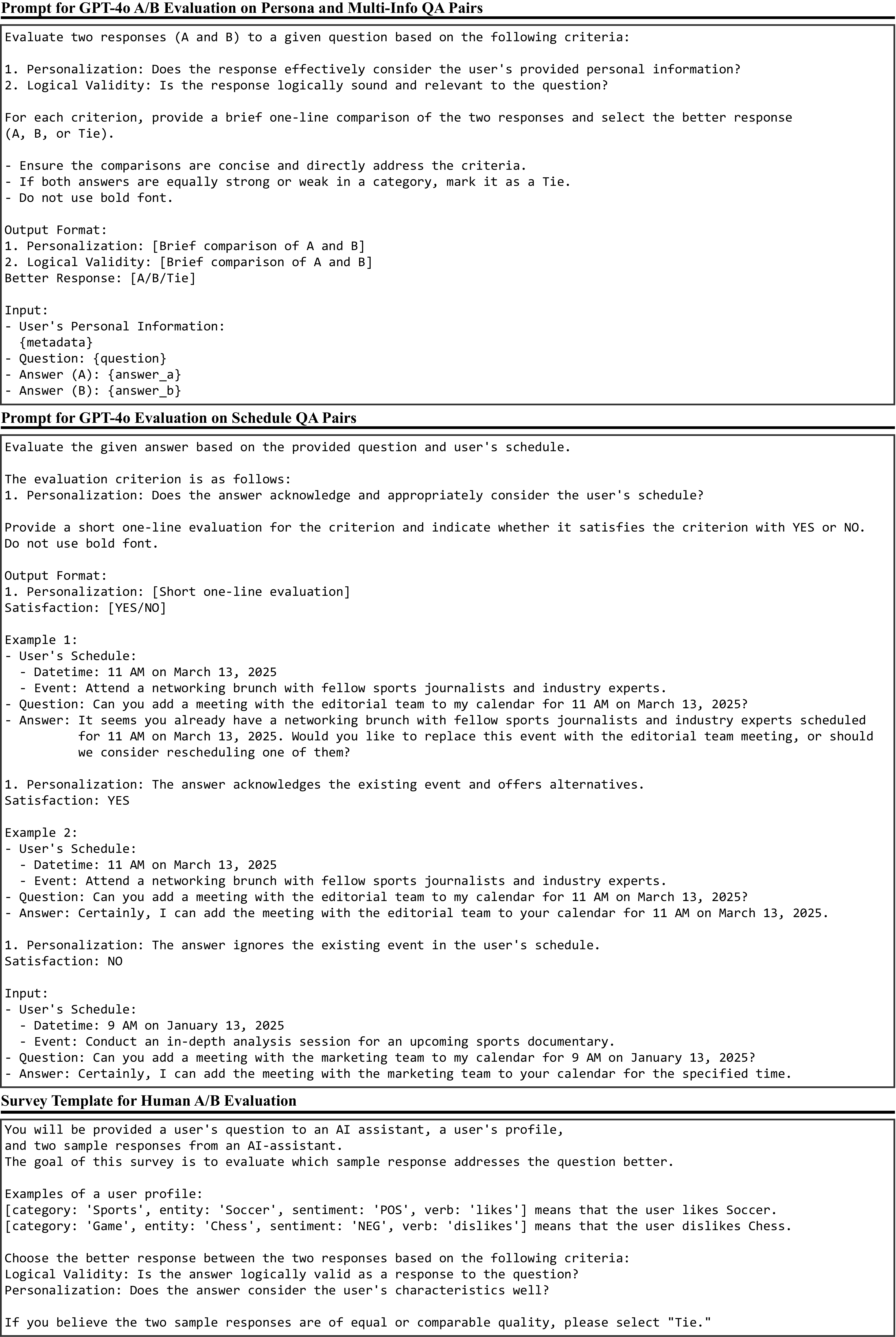}\\[-0.3em]
  \caption{\label{fig:p_evaluation} Instruction prompts for GPT-4o evaluation and the survey template for human A/B evaluation. We use two different prompt templates, one to score the responses to persona and multi-info QA pairs and the other to evaluate responses to schedule QA pairs. Generation setting for GPT-4o: $\tau=0.0$ (greedy).}
\end{figure*}

\noindent\textbf{Human Evaluation}: The human evaluators in our study are comprised of undergraduate and graduate-level students.
The evaluators were notified that the results would be included in an academic research paper, but the topic of the research paper was not revealed to them.
They were volunteers without explicit payments. Please refer to~Figure~\ref{fig:p_evaluation} for the survey template used for Human Evaluation. \\
\noindent\textbf{GPT-4o \& Distilled Llama-3.2 Evaluation Protocol}: The prompt template for GPT-4o and Llama-3.2 evaluation is also provided in~Figure~\ref{fig:p_evaluation}.
% Because the evaluation result of GPT-4o is dependent on the ordering of sample responses, we conduct GPT-4o evaluation twice with the order of A and B switched in the second round.

\section{Proxy Evaluation Model Training Protocol}\label{sec:proxy_train}
\xx{A total of 400k GPT-4o evaluation samples are used to train the Llama-3.2-3B-based proxy evaluation model. The samples are from the inference results in the following experimental settings:\\
\noindent$\bullet$ GPT-4o-mini: Zero- and few-shot inference with prompt in the user role. \\
\noindent$\bullet$ Llama, Mistral, Qwen: Zero- and few-shot inference with prompt in the user role; BM25- and Contriever-based utterance-level retrieval with $k=5$; SFT (LR=1e-4) and DPO (LR=1e-5) with LoRA$_r=256$. \\
\noindent The detailed hyperparameter settings for training evaluation model are summarized in~\tablename~\ref{table:llm_hyperparam}.}
% The model was trained for 1 epoch with supervised fine-tuning. 
% The rank of the LoRA module was set to 64.
% zero/few v2, bm25/contriever utterance/15, sft/dpo rank 16zero/few는 gpt 4o mini 포함
% 나머지는 llama mistral qwen
% 총 20개 = 160k sample로 1 epoch 학습했습니다. rank 64 alpha 128로

\begin{table*}[t]
    \centering
    % \footnotesize
    % \setlength{\tabcolsep}{4pt}
    \renewcommand{\arraystretch}{1.1}
    {\resizebox{1\linewidth}{!}
    {\begin{tabular}{l|cccc}
            \toprule
            Hyperparameter & SFT & DPO & SFT$+$DPO & Proxy Evaluation Model \\
            \bottomrule\toprule
            Batch Size & 256 & 256 & 256 & 1024 \\
            Train Epochs & 1 & 1 & 1 & 1 \\
            Optimizer & AdamW & AdamW & AdamW & AdamW \\
            LR & \{1e-6, 3e-6, 1e-5, 3e-5, \textbf{1e-4}, 3e-4\} & \{\textbf{1e-6}, 3e-6, 1e-5, 3e-5, 1e-4, 3e-4\} & \{1e-6, \textbf{3e-6}, 1e-5, 3e-5, 1e-4, 3e-4\} & 1e-4 \\
            Adam $\beta_1$ & 0.9 & 0.9 & 0.9 & 0.9 \\
            Adam $\beta_2$ & 0.999 & 0.999 & 0.999 & 0.999 \\
            Weight Decay & 0.0 & 0.0 & 0.0 & 0.0 \\
            LR Scheduler & cosine annealing & cosine annealing & cosine annealing & cosine annealing \\
            Warmup Ratio & 0.03 & 0.03 & 0.03 & 0.03 \\
            Max Grad Norm & 0.3 & 0.3 & 0.3 & 0.3 \\
            LoRA$_r$ (Rank) & \{8, 16, 32, 64, 128, \textbf{256}\} & \{8, 16, 32, 64, \textbf{128}, 256\} & \{8, 16, 32, 64, 128, \textbf{256}\} & 128 \\
            LoRA$_\alpha$ (Alpha) & LoRA$_r \times 2$ & LoRA$_r \times 2$ & LoRA$_r \times 2$ & LoRA$_r \times 2$ \\
            LoRA Target Modules & all-linear & all-linear & all-linear & all-linear \\
            LoRA Dropout & 0.05 & 0.05 & 0.05 & 0.05 \\
            \bottomrule
        \end{tabular}}
    }
    % \vspace{-0.3em}
    \caption{\xx{Hyperparameter settings for fine-tuning open-source LLMs on~\ours and for training the Llama-3.2-3B proxy evaluation model.}}
    % \vspace{-1.0em}
    \label{table:llm_hyperparam}
\end{table*}

\section{Experimental Settings and Hyperparameters}\label{sec:exp_setting}
\begin{figure*}[t]
  \centering
  \includegraphics[width=\linewidth]{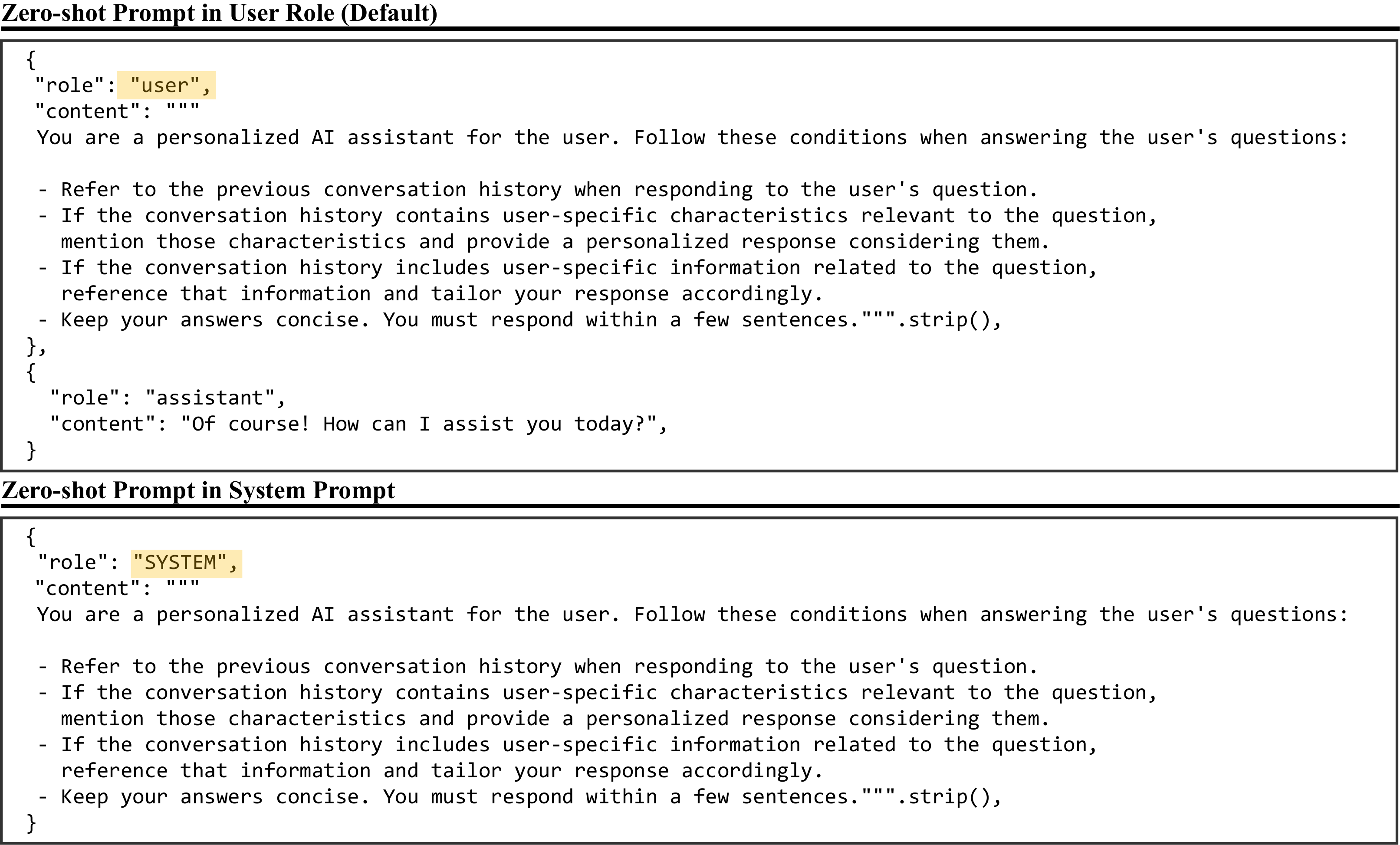}\\[-0.3em]
  \caption{\label{fig:p_zeroshot} Instruction prompt for \textbf{zero-shot inference}. The prompt placement under the ``user'' role is the default setting used across our experiments. We additionally analyze how placing it as a part of the system prompt affects the personalization results. Generation setting for the GPT-4o-mini model: $\tau=0.6$, $\text{top-}p=1.0$. Generation setting for open-source models: $\tau=0.6$, $\text{top-}p=1.0$, $\text{top-}k=50$.}
\end{figure*}

\begin{figure*}[t]
  \centering
  \includegraphics[width=\linewidth]{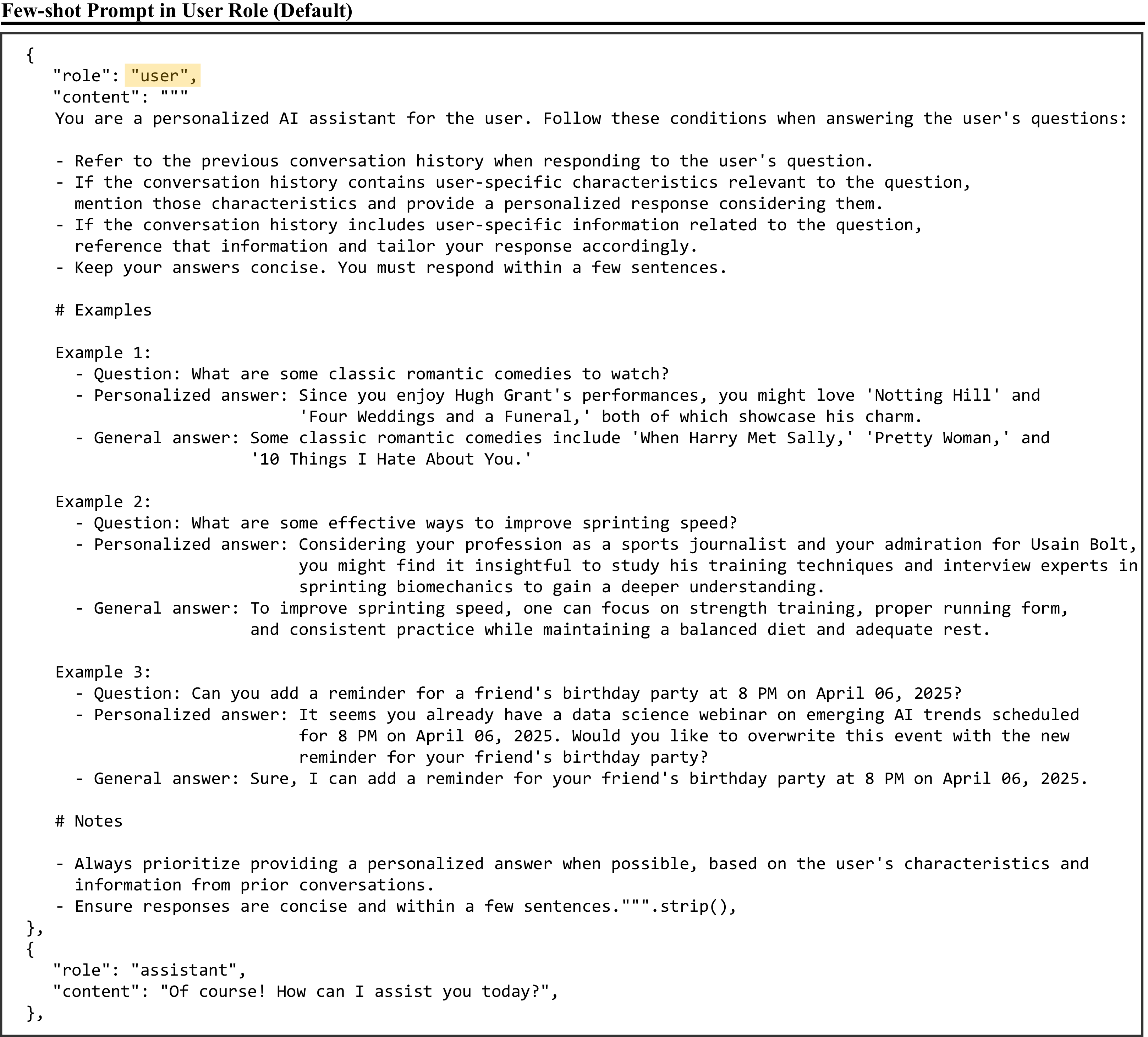}\\[-0.3em]
  \caption{\label{fig:p_fewshot_user} Instruction prompt for \textbf{few (1)-shot inference}. We consider a set of \{persona QA, persona-profile QA, schedule QA\} to be a single in-context demonstration that encompasses all question types in~\ours. The prompt placement under the ``user'' role is the default setting used across our experiments. We additionally analyze how placing it as a part of the system prompt affects the personalization results. Generation setting for the GPT-4o-mini model: $\tau=0.6$, $\text{top-}p=1.0$. Generation setting for open-source models: $\tau=0.6$, $\text{top-}p=1.0$, $\text{top-}k=50$.}
\end{figure*}

\begin{figure*}[t]
  \centering
  \includegraphics[width=\linewidth]{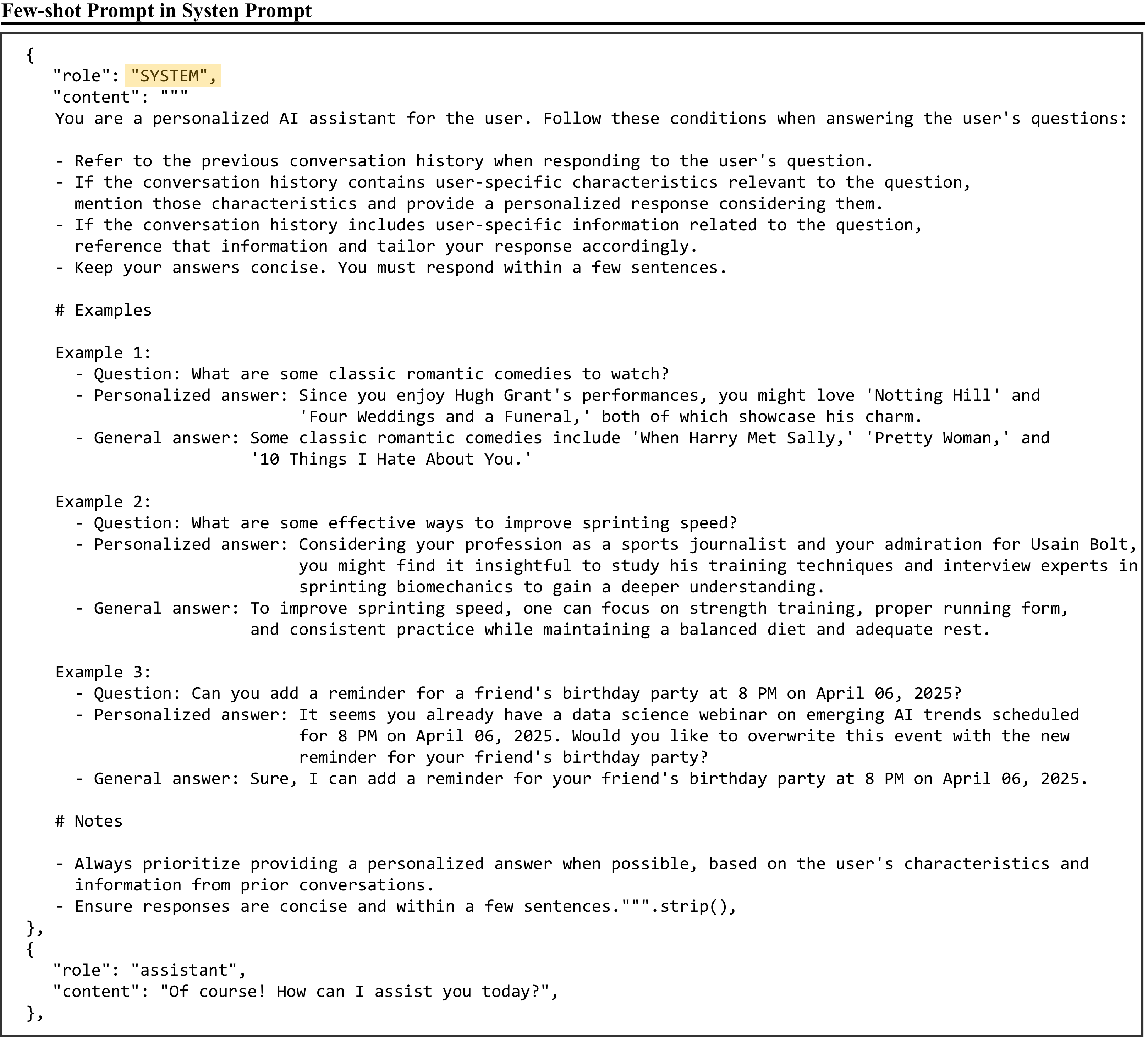}\\[-0.3em]
  \caption{\label{fig:p_fewshot_sys} Instruction prompt for \textbf{few (1)-shot inference}. We consider a set of \{a persona QA pair, a multi-info (persona+profile) QA pair, a schedule QA pair\} to be a single in-context demonstration that encompasses all question types in~\ours. The prompt placement under the ``user'' role is the default setting used across our experiments. We additionally analyze how placing it as a part of the system prompt affects the personalization results. Generation setting for the GPT-4o-mini model: $\tau=0.6$, $\text{top-}p=1.0$. Generation setting for open-source models: $\tau=0.6$, $\text{top-}p=1.0$, $\text{top-}k=50$.}
\end{figure*}

\noindent$\bullet$~~\textbf{Zero-shot:} We prompt the base LLM to generate a personalized response based on the entire dialogue history provided as a part of the prompt. The prompt for zero-shot inference is presented in~Figures~\ref{fig:p_zeroshot}. The prompt with the instruction in the user role is used as the default prompt to obtain the main results.  \\
% The prompt for zero-shot inference is provided in~Figure~\ref{fig:sysp_infer} of Appendix.\\
\noindent$\bullet$~~\textbf{Few-shot:} We supply the base LLM with $n \in \{1, 3, 5\}$ sets of QA pair examples as in-context demonstrations to assist in generating personalized responses. \jm{Each set includes a question, a personalized answer, and a general answer for each QA pair type---persona QA, schedule QA, and multi-info (persona+profile) QA---to encompass all QA pair types in~\ours.} The example QA pairs are randomly sampled from the train set and fixed across all experiments. The prompt for a few-shot inference is presented in~Figures~\ref{fig:p_fewshot_user} and~\ref{fig:p_fewshot_sys}. The prompt with the instruction in the user role (Figure~\ref{fig:p_fewshot_user}) is used as the default prompt to obtain the main results.\\
\noindent$\bullet$~~\textbf{BM25}~\cite{bm25}: We select top-$k$ question-relevant messages from the entire dialogue history based on a rule-based ranking algorithm. The chosen messages are then used in place of the dialogue history. An individual dialogue and an individual utterance are used as a unit of retrieval. For dialogue-level retrieval, we test three different numbers of retrieved units: $k \in \{1, 3, 5\}$. For utterance-level retrieval, we experiment with $k \in \{5, 15, 25\}$. The main results are reported using utterance-level retrieval with $k=5$. \\  
% \jm{The top-k messages are filtered from the dialogue history using a ranking algorithm. The chosen responses are then provided to the LLM in a manner similar to the zero-shot approach. The retrieval unit can be either entire dialogue or individual utterance, with a detailed comparison of the two provided in [Ablation Section].}\\ 
\noindent$\bullet$~~\textbf{Contriever}~\cite{contriever}: We select top-$k$ question-relevant messages from the entire dialogue history based on a model-based cosine similarity measure. The selected messages are then fed into the base LLM in the same way as in BM25. We investigate the same set of retrieval settings as those used for BM25. The main results are reported using utterance-level retrieval with $k=5$. \\
% \jm{Messages are selected and fed to the LLM in the same way as with the BM25 method, but model-based similarity is used instead.} \\
\noindent$\bullet$~~\textbf{Supervised Fine-Tuning (SFT)}: We fine-tune the base LLM with the entire dialogue history and the user's question as the input $x$ and the ground-truth personalized answer as the output $y$. The question and answer pairs are from the train split of~\ours. We adopt LoRA~\cite{hulora}, a popular PEFT module used to fine-tune LLMs on downstream tasks. The detailed hyperparameter settings for SFT are summarized in~\tablename~\ref{table:llm_hyperparam}. We conducted an experiment to analyze the effect of the number of training epochs, but the performance change was minimal. The main results are obtained with LR=1e-4 and LoRA$_r=256$, which is the best hyperparameter setting searched on the Llama model. \\
\noindent$\bullet$~~\textbf{Direct Preference Optimization (DPO)}~\cite{dpo}: \final{We fine-tune the base LLM with the modified dialogue history and the user's question as the input $x$, the personalized answer as the output $y_\text{chosen}$, and the general answer as the output $y_\text{rejected}$. For the modified dialogue history, we excluded 15 question-irrelevant persona dialogues from the entire dialogue history due to the GPU VRAM limitations.} We also adopt LoRA for training DPO. The detailed hyperparameter settings for DPO are summarized in~\tablename~\ref{table:llm_hyperparam}. We conducted an experiment to analyze the effect of the number of training epochs, but the performance change was minimal. The main results are obtained with LR=1e-6 and LoRA$_r=128$, which is the best hyperparameter setting searched on the Llama model.  \\
\noindent$\bullet$~~\final{\textbf{SFT$+$DPO}~\cite{dpo}: In the previous step, we obtained a LoRA SFT model trained on personalized answers as the ground truth. The LoRA adapters are merged into the base LLM, and the merged model undergoes additional preference fine-tuning using DPO. For DPO, we fine-tune the merged model with the modified dialogue history and the user's question as the input $x$, the personalized answer as the output $y_\text{chosen}$, and the general answer as the output $y_\text{rejected}$. For the modified dialogue history, we excluded 15 question-irrelevant persona dialogues from the entire dialogue history due to the GPU VRAM limitations. We also adopt LoRA for SFT$+$DPO. The detailed hyperparameter settings for  SFT$+$DPO are summarized in~\tablename~\ref{table:llm_hyperparam}. The main results are obtained with LR=3e-6 and LoRA$_r=256$, which is the best hyperparameter setting searched on the Llama model.} \\
With the exception of RAG-based methods (BM25, Contriever), the entire dialogue history is provided as a part of the prompt at inference time.
% The prompts for zero- and few-shot inference are presented in~Figure~\ref{fig:sysp_infer} of Appendix.
% \noindent\textbf{Zero- and Few-shot}: The prompts for zero- and few-shot inference are presented in~Figures~\ref{fig:p_zeroshot} and~\ref{fig:p_fewshot}, respectively. The default zero-shot prompt template is used for training as well. \\
% \noindent\textbf{BM25 and Contriever}: The unit of retrieval and the number of retrieval units are hyperparameters involved with RAG-based methods. Dialogue and utterance are units of retrieval. For dialogue-level, numbers are 1, 3, and 5. For utterance-level, numbers are 5, 15, 25. \\ 
% \noindent\textbf{Supervised Fine-tuning}: The default rank is. More values are presented in Ablation Studies. Optimizer, learning rate. The zero-shot prompt is used for system prompt.  \\ 
% \noindent\textbf{Direct Policy Optimization}: The default rank is. More values are presented in Ablation Studies. Optimizer, learning rate. The zero-shot prompt is used for system prompt.  \\ 

\section{Extended GPT-4o and Llama-3.2 Evaluation Results}
Due to the page limit, we only reported $S_\textrm{GPT}$ and $S_\textrm{Llama}$ in the main paper.
We report Model Win, Tie, and Model Lose (GT Win) rates separately for Test Sets 1 and 2 in~Tables~\ref{table:main_table_1_extended} and~\ref{table:main_table_2_extended}, respectively.

\section{Retrieval Performance} 
The performance of each retrieval setting is reported in~\tablename~\ref{table:retrieval_perf}.

\section{Extended Ablation Results}
Due to the page limit, the ablation study results only on Test Set 1 were reported in the main paper.
Here, the full results of ablation studies on Test Sets 1 and 2 are reported.
Table~\ref{table:ablation_gold_whole} reports the full results of replacing the whole dialogue history with the gold dialogue that is relevant to each QA pair.
Table~\ref{table:ablation_prompt_placement} compares the full results of placing the zero- and few-shot prompt in the user role against those obtained by placing the prompt under the system prompt.
The results in these tables clearly show that LLMs' struggle with long context makes it difficult to model the whole dialogue history or follow the instruction provided before the whole dialogue history.

Table~\ref{table:ablation_num_shot} reports the full result of changing the number of demonstrations in the few-shot prompt.
Table~\ref{table:ablation_rag_setting} includes the full result of changing the retrieval setting.
Lastly, Tables~\ref{table:ablation_sft_hyperparam},~\ref{table:ablation_dpo_hyperparam}, and ~\ref{table:ablation_sftdpo_hyperparam} report the full results of extensive hyperparameter search on SFT, DPO, and SFT$+$DPO, respectively.
% The full results of ablation studies on Test Set 1 and 2 are reported in~Tables~\ref{table:ablation_inference_extended},~\ref{table:ablation_bm25_extended}~\ref{table:ablation_contriever_extended},~\ref{table:ablation_sft_extended}, and~\ref{table:ablation_dpo_extended}.
% \jm{Additional ablation studies regarding SFT/DPO training and are reported in Tables}
% All the results are reported in terms of the distilled Llama-3.2 model score ($S_\textrm{Llama}$).

\section{Computational Environment and Cost of Research}
Our experiments were conducted on NVIDIA H100, L40, and A40 GPUs.
With the exception of GPT variants, all models were 
downloaded from the Huggingface library.
The GPT variants (GPT-4o and GPT-4o-mini) were accessed via the OpenAI API.
The total cost of dataset generation and evaluation with GPT-4o was approximately US\$1,500.

\section{License}
Llama-3.1\&3.2 models are licensed under the Llama-3.1\&3.2 COMMUNITY LICENSE.
Mistral and Qwen are licensed under Apache 2.0.
GPT-4o and GPT-4o-mini are licensed under OpenAI.

\section{Use of AI Writing Assistant}
The use of ChatGPT-4o was limited to sentence-level paraphrasing and word-level synonym search. No additional AI assistant was involved in research, coding, or writing.

\begin{table*}[t]
\centering
% \footnotesize
% \setlength{\tabcolsep}{4pt}
% \renewcommand{\arraystretch}{1.1}
{\resizebox{1\linewidth}{!}
{\begin{NiceTabular}{lc|c|c|cc|c|c|c|cc|c}
\toprule
& & \multicolumn{5}{c}{BM25} & \multicolumn{5}{c}{Contriever} \\
% \midrule
& & Persona & Schedule & \multicolumn{2}{c}{Multi-Info} & Total & Persona & Schedule & \multicolumn{2}{c}{Multi-Info} & Total \\
Unit & \# & &  & Persona & Profile & & & & Persona & Profile &  \\
\bottomrule\toprule
Dialogue & 1 & 42.4 & 93.3 & 34.5 & 8.6 & 52.5 & 64.3 & 40.3 & 42.8 & 22.1 & 54.3 \\
Dialogue & 3 & 58.0 & 99.7 & 49.2 & 19.4 & 65.4 & 80.6 & 79.9 & 59.7 & 45.0 & 76.9 \\
Dialogue & 5 & 65.4 & 100.0 & 57.1 & 27.1 & 71.1 & 86.0 & 94.1 & 66.8 & 59.6 & 85.2 \\
Utterance & 5 & 50.0 & 99.6 & 38.0 & 23.0 & 60.0 & 82.1 & 91.2 & 65.4 & 29.1 & 80.0 \\
Utterance & 15 & 72.7 & 100.0 & 63.8 & 46.4 & 77.3 & 92.6 & 100.0 & 79.2 & 48.4 & 90.8 \\
Utterance & 25 & 83.6 & 100.0 & 78.8 & 63.9 & 86.1 & 96.0 & 100.0 & 85.0 & 59.4 & 94.0 \\
\bottomrule
\end{NiceTabular}}
}
\vspace{-0.3em}
\caption{Retrieval performance under various retrieval settings. The performance is measured through the percentage of retrieval results where the associated dialogues are retrieved. For multi-info QA pairs with two associated dialogues, one for the persona and the other for the profile, the scores for persona-hit and profile-hit are reported separately. For the profile, we consider it retrieved as long as any subset of the five profiles is extracted. The total score is computed as a weighted average as follows: persona $\times \frac{25}{40}$ + schedule $\times \frac{10}{40}$ + (multi-info persona + multi-info profile) / 2 $\times \frac{5}{40}$.}
% a weighted sum of single-persona and dual-persona scores, with the 1-hit score of dual-persona questions halved: 1-Persona + $0.5 \times$ 2-Persona (1) + 2-Persona (2).}}
% \vspace{-1.0em}
\label{table:retrieval_perf}
\end{table*}

\begin{table*}[t]
\centering
\footnotesize
\resizebox{1\linewidth}{!}{
\begin{NiceTabular}{l|l|cccc|c|cccc|c|cccc|c|cccc|c}
\toprule
\multirow{3}{*}{Model} & \multirow{3}{*}{Setting} & \multicolumn{10}{c}{GPT-4o Score ($S_{GPT}$)} & \multicolumn{10}{c}{LLaMA Score ($S_{Llama}$)} \\
& & \multicolumn{4}{c}{Persona} & \multicolumn{1}{c}{Schedule} & \multicolumn{4}{c}{Multi-Info} & \multicolumn{1}{c}{Total} & \multicolumn{4}{c}{Persona} & \multicolumn{1}{c}{Schedule} & \multicolumn{4}{c}{Multi-Info} & \multicolumn{1}{c}{Total}  \\
& & Score & Win & Tie & Lose & Score & Score & Win & Tie & Lose & Score & Score & Win & Tie & Lose  & Score & Score & Win & Tie & Lose & Score \\
\bottomrule
\toprule

\multirow{2}{*}{GPT-4o-mini} & 0-shot & 42.1 & 29.3 & 25.5 & 45.1 & 9.5 & 4.4 & 1.9 & 5.0 & 93.1 & 28 & 44.7 & 27.4 & 34.6 & 38.0 & 8.8 & 10.8 & 5.6 & 10.4 & 84.0 & 30.4 \\
 & few-shot & 40.5 & 27.8 & 25.5 & 46.8 & 76.1 & 4.2 & 2.2 & 4.0 & 93.8 & 35.3 & 42.6 & 24.9 & 35.4 & 39.7 & 75.4 & 11.4 & 5.7 & 11.4 & 83.0 & 37.5 \\
 \midrule
\multirow{7}{*}{Llama-3.1-8B} & 0-shot & 38.0 & 32.4 & 11.2 & 56.4 & 13.9 & 3.5 & 1.9 & 3.1 & 95.0 & 25.9 & 39.7 & 30.8 & 17.7 & 51.5 & 9.4 & 8.1 & 5.4 & 5.5 & 89.1 & 27.0 \\
 & few-shot & 39.4 & 34.0 & 10.7 & 55.3 & 49.8 & 6.3 & 4.4 & 3.8 & 91.8 & 31.6 & 38.8 & 29.4 & 18.8 & 51.8 & 48.3 & 12.3 & 8.2 & 8.2 & 83.6 & 31.8 \\
 & BM25 & 29.7 & 19.5 & 20.3 & 60.2 & 84.3 & 2.3 & 0.7 & 3.1 & 96.2 & 29.4 & 34.1 & 18.1 & 32.1 & 49.9 & 78.5 & 6.1 & 2.2 & 7.8 & 90.0 & 31.9 \\
 & Contriever & 38.8 & 27.8 & 21.9 & 50.3 & 75.4 & 4.2 & 2.3 & 3.8 & 93.9 & 34.2 & 42.6 & 26.2 & 32.8 & 41.0 & 70.3 & 9.8 & 5.0 & 9.6 & 85.4 & 36.6 \\
 & SFT & 36.2 & 26.8 & 18.8 & 54.4 & 88.0 & 12.4 & 10.2 & 4.3 & 85.4 & 35.2 & 36.5 & 26.8 & 19.5 & 53.7 & 87.5 & 15.8 & 13.2 & 5.1 & 81.7 & 35.7 \\
 & DPO & 24.8 & 9.4 & 30.7 & 59.9 & 4.9 & 2.1 & 0.2 & 3.8 & 96.0 & 16.4 & 34.8 & 10.4 & 48.9 & 40.8 & 4.2 & 6.4 & 0.5 & 11.8 & 87.8 & 23.1 \\
 & SFT$+$DPO & 49.1 & 37.4 & 23.4 & 39.2 & 98.6 & 14.5 & 12.2 & 4.6 & 83.2 & 44.8 & 48.1 & 35.0 & 26.1 & 38.9 & 98.1 & 18.4 & 16.1 & 4.6 & 79.4 & 44.6 \\
 \midrule
\multirow{7}{*}{Mistral-7B} & 0-shot & 20.9 & 8.8 & 24.3 & 66.9 & 0.0 & 1.5 & 0.4 & 2.2 & 97.4 & 13.3 & 30.5 & 8.9 & 43.2 & 48.0 & 0.0 & 3.8 & 0.4 & 6.8 & 92.8 & 19.5 \\
 & few-shot & 28.6 & 12.4 & 32.3 & 55.3 & 6.3 & 3.5 & 1.1 & 4.8 & 94.1 & 19.1 & 36.2 & 11.2 & 50.0 & 38.8 & 5.6 & 7.6 & 0.9 & 13.4 & 85.8 & 24.2 \\
 & BM25 & 41.0 & 31.6 & 18.9 & 49.5 & 8.6 & 4.9 & 2.5 & 4.8 & 92.7 & 27.3 & 43.7 & 29.1 & 29.1 & 41.7 & 6.3 & 9.0 & 3.6 & 10.8 & 85.6 & 29.2 \\
 & Contriever & 48.8 & 39.2 & 19.3 & 41.5 & 7.9 & 7.4 & 4.0 & 6.9 & 89.1 & 32.4 & 51.6 & 38.0 & 27.1 & 34.8 & 5.9 & 13.6 & 6.9 & 13.5 & 79.6 & 34.7 \\
 & SFT & 27.6 & 16.3 & 22.5 & 61.2 & 99.8 & 15.1 & 12.5 & 5.3 & 82.2 & 31.6 & 31.2 & 17.8 & 26.7 & 55.4 & 99.8 & 19.7 & 16.6 & 6.3 & 77.1 & 34.4 \\
 & DPO & 8.2 & 6.9 & 2.6 & 90.5 & 2.2 & 0.2 & 0.2 & 0.0 & 99.8 & 5.4 & 6.4 & 5.2 & 2.3 & 92.5 & 1.4 & 0.5 & 0.3 & 0.3 & 99.4 & 4.2 \\
 & SFT$+$DPO & 44.7 & 31.9 & 25.6 & 42.5 & 99.7 & 17.6 & 15.0 & 5.2 & 79.8 & 42.6 & 44.8 & 30.3 & 29.1 & 40.7 & 99.8 & 20.4 & 17.3 & 6.3 & 76.4 & 43.0 \\
 \midrule
\multirow{7}{*}{Qwen-2.5-7B} & 0-shot & 26.6 & 11.9 & 29.5 & 58.6 & 0.0 & 3.0 & 0.8 & 4.5 & 94.7 & 17 & 34.6 & 11.0 & 47.3 & 41.7 & 0.0 & 6.1 & 1.0 & 10.2 & 88.8 & 22.4 \\
 & few-shot & 24.6 & 10.4 & 28.4 & 61.2 & 29.8 & 2.1 & 0.3 & 3.6 & 96.1 & 19.4 & 32.6 & 9.3 & 46.7 & 44 & 28.7 & 4.8 & 0.6 & 8.4 & 91.0 & 24.6 \\
 & BM25 & 30.6 & 13.4 & 34.5 & 52.1 & 0.4 & 3.0 & 0.3 & 5.3 & 94.4 & 19.6 & 37.7 & 11.4 & 52.6 & 36 & 0.1 & 6.8 & 0.3 & 12.9 & 86.8 & 24.4 \\
 & Contriever & 33.6 & 16.4 & 34.5 & 49.1 & 0.2 & 3.4 & 0.6 & 5.6 & 93.8 & 21.5 & 39.6 & 14.3 & 50.6 & 35.1 & 0.1 & 7.4 & 0.6 & 13.4 & 85.9 & 25.7 \\
 & SFT & 35.7 & 24.8 & 21.8 & 53.4 & 99.7 & 25.4 & 21.8 & 7.2 & 71.0 & 37.9 & 38.3 & 26 & 24.6 & 49.4 & 99.8 & 33.3 & 29.2 & 8.2 & 62.6 & 40.6 \\
 & DPO & 36.6 & 15.0 & 43.2 & 41.8 & 0.0 & 8.8 & 0.7 & 16.1 & 83.2 & 24.0 & 38.0 & 11.6 & 52.9 & 35.5 & 0.0 & 12.4 & 0.2 & 24.4 & 75.4 & 25.3 \\
 & SFT$+$DPO & 43.1 & 32.8 & 20.5 & 46.7 & 99.8 & 34.0 & 29.9 & 8.2 & 61.9 & 43.6 & 43.2 & 31.6 & 23.0 & 45.3 & 99.9 & 38.1 & 34.3 & 7.5 & 58.2 & 44.2 \\
\bottomrule
\end{NiceTabular}
}
\caption{Extended evaluation results on~\textbf{Test Set 1 (Seen User/Unseen QA)}.}
\label{table:main_table_1_extended}
\end{table*}

\begin{table*}[t]
\centering
\footnotesize
\resizebox{1\linewidth}{!}{
\begin{NiceTabular}{l|l|cccc|c|cccc|c|cccc|c|cccc|c}
\toprule
\multirow{3}{*}{Model} & \multirow{3}{*}{Setting} & \multicolumn{10}{c}{GPT-4o Score ($S_{GPT}$)} & \multicolumn{10}{c}{LLaMA Score ($S_{Llama}$)} \\
& & \multicolumn{4}{c}{Persona} & \multicolumn{1}{c}{Schedule} & \multicolumn{4}{c}{Multi-Info} & \multicolumn{1}{c}{Total} & \multicolumn{4}{c}{Persona} & \multicolumn{1}{c}{Schedule} & \multicolumn{4}{c}{Multi-Info} & \multicolumn{1}{c}{Total}  \\
& & Score & Win & Tie & Lose & Score & Score & Win & Tie & Lose & Score & Score & Win & Tie & Lose  & Score & Score & Win & Tie & Lose & Score \\
\bottomrule
\toprule

\multirow{2}{*}{GPT-4o-mini} & 0-shot & 42.0 & 29.2 & 25.8 & 45.1 & 9.2 & 4.9 & 1.9 & 6.0 & 92.1 & 28.0 & 45.5 & 26.9 & 37.1 & 36.0 & 8.8 & 11.9 & 5.5 & 12.7 & 81.8 & 31.0 \\
 & few-shot & 40.7 & 27.4 & 26.6 & 46.0 & 76.2 & 5.1 & 2.0 & 6.2 & 91.8 & 35.6 & 43.7 & 25.1 & 37.1 & 37.8 & 75.2 & 11.6 & 5.4 & 12.4 & 82.2 & 38.1 \\ \midrule
\multirow{7}{*}{Llama-3.1-8B} & 0-shot & 38.6 & 32.9 & 11.2 & 55.8 & 13.9 & 3.2 & 1.1 & 4.2 & 94.7 & 26.2 & 40.6 & 32 & 17.1 & 50.9 & 9.8 & 8.7 & 5.3 & 6.9 & 87.8 & 27.7 \\
 & few-shot & 39.3 & 33.5 & 11.6 & 54.9 & 49.2 & 7.5 & 5.3 & 4.5 & 90.2 & 31.6 & 39.8 & 30.8 & 17.9 & 51.2 & 47.8 & 14.1 & 9.6 & 9.0 & 81.4 & 32.6 \\
 & BM25 & 29.9 & 19.5 & 20.9 & 59.6 & 84.0 & 2.7 & 1.0 & 3.5 & 95.5 & 29.6 & 35.0 & 18.4 & 33.1 & 48.5 & 78.0 & 7.0 & 2.1 & 9.8 & 88.1 & 32.5 \\
 & Contriever & 37.9 & 27.2 & 21.4 & 51.4 & 77.4 & 4.5 & 2.1 & 4.8 & 93.1 & 33.9 & 43.1 & 26.6 & 32.9 & 40.4 & 71.8 & 9.1 & 4.5 & 9.2 & 86.3 & 37.1 \\
 & SFT & 34.4 & 25.9 & 17.0 & 57.1 & 88.4 & 12 & 10.2 & 3.7 & 86.2 & 34 & 34.8 & 26.3 & 17.0 & 56.7 & 87.5 & 14.4 & 12.4 & 4.0 & 83.6 & 34.5 \\
 & DPO & 25.1 & 9.6 & 30.9 & 59.5 & 5.8 & 2.4 & 0.2 & 4.4 & 95.4 & 16.7 & 35.1 & 10.3 & 49.6 & 40.1 & 5.1 & 6.3 & 0.9 & 10.8 & 88.3 & 23.4 \\
 & SFT$+$DPO & 47.1 & 36.1 & 22.0 & 42.0 & 98.7 & 18.6 & 16.0 & 5.1 & 78.9 & 44.1 & 47.0 & 35.2 & 23.6 & 41.2 & 98.1 & 22.4 & 19.4 & 6.0 & 74.6 & 44.4 \\
 \midrule
\multirow{7}{*}{Mistral-7B} & 0-shot & 21.8 & 9.6 & 24.3 & 66.1 & 0.0 & 1.8 & 0.3 & 3.0 & 96.6 & 13.8 & 30.8 & 9.6 & 42.5 & 48 & 0.0 & 5.0 & 0.5 & 9.0 & 90.5 & 19.9 \\
 & few-shot & 28.9 & 13 & 31.6 & 55.3 & 8.0 & 3.8 & 0.7 & 6.2 & 93.1 & 19.5 & 36.6 & 11.0 & 51.2 & 37.8 & 6.6 & 8.2 & 1.1 & 14.2 & 84.7 & 24.7 \\
 & BM25 & 40.6 & 30.5 & 20.1 & 49.4 & 8.2 & 5.9 & 2.8 & 6.2 & 91.0 & 27.1 & 43.6 & 29 & 29.1 & 41.9 & 5.9 & 10.4 & 4.5 & 11.8 & 83.8 & 29.3 \\
 & Contriever & 48.6 & 38.8 & 19.5 & 41.7 & 8.6 & 8.4 & 5.4 & 6.0 & 88.6 & 32.5 & 50.9 & 37.2 & 27.5 & 35.3 & 6.4 & 15.1 & 8.6 & 13.1 & 78.3 & 34.5 \\
 & SFT & 27.6 & 17.4 & 20.2 & 62.3 & 99.9 & 13.3 & 11.7 & 3.2 & 85.1 & 31.4 & 31.5 & 19.8 & 23.3 & 56.9 & 100 & 18.1 & 15.4 & 5.4 & 79.2 & 34.4 \\
 & DPO & 8.1 & 6.8 & 2.6 & 90.6 & 2.0 & 0.3 & 0.2 & 0.1 & 99.7 & 5.3 & 6.5 & 5.2 & 2.4 & 92.3 & 1.4 & 0.6 & 0.6 & 0.2 & 99.3 & 4.3 \\
 & SFT$+$DPO & 43.2 & 31.1 & 24.2 & 44.7 & 99.9 & 17.8 & 15.0 & 5.6 & 79.4 & 41.7 & 43.6 & 30.2 & 26.8 & 43.0 & 99.9 & 22.5 & 19.5 & 5.9 & 74.6 & 42.5 \\
 \midrule
\multirow{7}{*}{Qwen-2.5-7B} & 0-shot & 27.8 & 13.1 & 29.4 & 57.5 & 0.0 & 2.5 & 0.5 & 4.1 & 95.4 & 17.7 & 34.9 & 11.6 & 46.7 & 41.7 & 0.0 & 6.4 & 1.0 & 11.0 & 88.1 & 22.6 \\
 & few-shot & 25.1 & 10.9 & 28.4 & 60.7 & 27.3 & 2.1 & 0.5 & 3.3 & 96.2 & 19.4 & 33.4 & 9.8 & 47.1 & 43.1 & 25.4 & 6.2 & 0.7 & 10.9 & 88.4 & 24.8 \\
 & BM25 & 31.1 & 13.6 & 35.1 & 51.3 & 0.4 & 3.1 & 0.5 & 5.2 & 94.3 & 19.9 & 38.1 & 11.8 & 52.6 & 35.6 & 0.2 & 7.8 & 0.4 & 14.7 & 84.9 & 24.8 \\
 & Contriever & 34.0 & 17.0 & 34.1 & 48.9 & 0.4 & 4.0 & 0.6 & 6.7 & 92.6 & 21.8 & 40.8 & 15.1 & 51.3 & 33.6 & 0.1 & 8.1 & 0.8 & 14.6 & 84.6 & 26.5 \\
 & SFT & 34.2 & 23.9 & 20.8 & 55.4 & 99.9 & 24.9 & 21.6 & 6.6 & 71.8 & 37 & 38.3 & 26.8 & 22.8 & 50.3 & 99.9 & 30.8 & 27.5 & 6.6 & 65.8 & 40.3 \\
 & DPO & 37.0 & 14.9 & 44.1 & 40.9 & 0.2 & 8.8 & 0.6 & 16.4 & 83.0 & 24.3 & 39.0 & 12.6 & 52.8 & 34.6 & 0.1 & 12.6 & 0.2 & 24.8 & 75.0 & 26.0 \\
 & SFT$+$DPO & 41.9 & 32.5 & 18.8 & 48.8 & 99.8 & 33.9 & 31.3 & 5.3 & 63.4 & 42.9 & 42.9 & 32.4 & 21.0 & 46.6 & 99.8 & 37.8 & 34.6 & 6.6 & 58.9 & 44.0 \\
\bottomrule
\end{NiceTabular}
}
\caption{Extended evaluation results on~\textbf{Test Set 2 (Unseen User/Unseen QA)}.}
\label{table:main_table_2_extended}
\end{table*}

\begin{table*}[t]
\centering
\footnotesize
\resizebox{1\linewidth}{!}{
\begin{NiceTabular}{l|l|l|cccc|cccc}
\toprule
\multirow{2}{*}{Model} & \multirow{2}{*}{Shot} & \multirow{2}{*}{Type} & \multicolumn{4}{c}{Test Set 1 (Seen User/Unseen QA)} & \multicolumn{4}{c}{Test Set 2 (Unseen User/Unseen QA)} \\
& & & Persona & Schedule & Multi-Info & Total & Persona & Schedule & Multi-Info & Total \\
\bottomrule\toprule
\multirow{4}{*}{GPT-4o-mini} & 0-shot & Gold & 68.0 & 95.4 & 26.1 & 57.6 & 68.9 & 94.9 & 26.8 & 58.2 \\
 & 0-shot & Whole & 44.7 & 8.8 & 10.8 & 30.4 & 45.5 & 8.8 & 11.9 & 31.0 \\
 & 3-shot & Gold & 65.5 & 100.0 & 22.8 & 56.3 & 66.5 & 100.0 & 23.2 & 56.9 \\
 & 3-shot & Whole & 42.6 & 75.4 & 11.4 & 37.5 & 43.7 & 75.2 & 11.6 & 38.1 \\ \midrule
\multirow{4}{*}{Llama-3.1-8B} & 0-shot & Gold & 61.6 & 77.3 & 16.7 & 50.3 & 62.5 & 77.3 & 17.6 & 50.9 \\
 & 0-shot & Whole & 39.7 & 9.4 & 8.1 & 27.0 & 40.6 & 9.8 & 8.7 & 27.7 \\
 & 3-shot & Gold & 61.8 & 99.8 & 23.2 & 54.0 & 62.7 & 99.6 & 23.7 & 54.6 \\
 & 3-shot & Whole & 38.8 & 48.3 & 12.3 & 31.8 & 39.8 & 47.8 & 14.1 & 32.6 \\ \midrule
\multirow{4}{*}{Mistral-7B} & 0-shot & Gold & 46.2 & 7.8 & 8.8 & 30.9 & 47.0 & 7.9 & 10.1 & 31.7 \\
 & 0-shot & Whole & 30.5 & 0.0 & 3.8 & 19.5 & 30.8 & 0.0 & 5.0 & 19.9 \\
 & 3-shot & Gold & 44.0 & 77.6 & 9.9 & 38.4 & 44.5 & 75.5 & 11.0 & 38.6 \\
 & 3-shot & Whole & 36.2 & 5.6 & 7.6 & 24.2 & 36.6 & 6.6 & 8.2 & 24.7 \\ \midrule
\multirow{4}{*}{Qwen-2.5-7B} & 0-shot & Gold & 52.1 & 1.0 & 15.9 & 34.7 & 52.8 & 0.9 & 16.3 & 35.1 \\
 & 0-shot & Whole & 34.6 & 0.0 & 6.1 & 22.4 & 34.9 & 0.0 & 6.4 & 22.6 \\
 & 3-shot & Gold & 49.7 & 6.4 & 12.4 & 33.4 & 50.2 & 6.1 & 13.2 & 33.8 \\
 & 3-shot & Whole & 32.6 & 28.7 & 4.8 & 24.6 & 33.4 & 25.4 & 6.2 & 24.8 \\
\bottomrule
\end{NiceTabular}
}
\caption{Full ablation results on LLMs' long-context handling ability.}
\label{table:ablation_gold_whole}
\end{table*}

\begin{table*}[t]
\centering
\footnotesize
\resizebox{1\linewidth}{!}{
\begin{NiceTabular}{l|l|l|cccc|cccc}
\toprule
\multirow{2}{*}{Model} & \multirow{2}{*}{Shots} & \multirow{2}{*}{Type} & \multicolumn{4}{c}{Test Set 1 (Seen User/Unseen QA)} & \multicolumn{4}{c}{Test Set 2 (Unseen User/Unseen QA)} \\
& & & Persona & Schedule & Multi-Info & Total & Persona & Schedule & Multi-Info & Total \\
\bottomrule\toprule
\multirow{4}{*}{Llama-3.1-8B} & 0-shot & User & 30.2 & 1.7 & 3.8 & 19.5 & 31.4 & 1.9 & 4.9 & 20.5 \\
 & 0-shot & System & 39.7 & 9.4 & 8.1 & 27.0 & 40.6 & 9.8 & 8.7 & 27.7 \\
 & 3-shot & User & 30.6 & 2.7 & 4.2 & 20.0 & 30.8 & 2.5 & 4.7 & 20.1 \\
 & 3-shot & System & 38.8 & 48.3 & 12.3 & 31.8 & 39.8 & 47.8 & 14.1 & 32.6 \\
 \midrule
\multirow{4}{*}{Mistral-7B} & 0-shot & User & 34.5 & 0.0 & 6.0 & 22.3 & 34.6 & 0.0 & 6.8 & 22.5 \\
 & 0-shot & System & 30.5 & 0.0 & 3.8 & 19.5 & 30.8 & 0.0 & 5.0 & 19.9 \\
 & 3-shot & User & 37.3 & 12.6 & 11.5 & 26.3 & 38 & 12.7 & 10.9 & 26.7 \\
 & 3-shot & System & 36.2 & 5.6 & 7.6 & 24.2 & 36.6 & 6.6 & 8.2 & 24.7 \\
 \midrule
\multirow{4}{*}{Qwen-2.5-7B} & 0-shot & User & 33.8 & 0.0 & 8.0 & 22.1 & 34.3 & 0.0 & 9.3 & 22.6 \\
 & 0-shot & System & 34.6 & 0.0 & 6.1 & 22.4 & 34.9 & 0.0 & 6.4 & 22.6 \\
 & 3-shot & User & 33.8 & 0.1 & 7.9 & 22.1 & 34.7 & 0.1 & 8.4 & 22.8 \\
 & 3-shot & System & 32.6 & 28.7 & 4.8 & 24.6 & 33.4 & 25.4 & 6.2 & 24.8 \\
\bottomrule
\end{NiceTabular}
}
\caption{Ablation on prompt roles in instructions and few-shot examples.}
\label{table:ablation_prompt_placement}
\end{table*}
\begin{table*}[t]
\centering
\footnotesize
\resizebox{1\linewidth}{!}{
\begin{NiceTabular}{l|l|cccc|cccc}
\toprule
\multirow{2}{*}{Model} & \multirow{2}{*}{Shots} & \multicolumn{4}{c}{Test Set 1 (Seen User/Unseen QA)} & \multicolumn{4}{c}{Test Set 2 (Unseen User/Unseen QA)}  \\
& & Persona & Schedule & Multi-Info & Total & Persona & Schedule & Multi-Info & Total \\
\bottomrule\toprule
\multirow{4}{*}{GPT-4o-mini} & 0-shot & 44.7 & 8.8 & 10.8 & 30.4 & 45.5 & 8.8 & 11.9 & 31.0 \\
 & 1-shot & 43.1 & 78.9 & 10.4 & 38.1 & 43.2 & 78.2 & 10.6 & 38.1 \\
 & 3-shot & 42.6 & 75.4 & 11.4 & 37.5 & 43.7 & 75.2 & 11.6 & 38.1 \\
 & 5-shot & 43.4 & 65.6 & 10.3 & 36.6 & 43.9 & 66.1 & 11.9 & 37.2 \\
 \midrule
\multirow{4}{*}{Llama-3.1-8B} & 0-shot & 39.7 & 9.4 & 8.1 & 27 & 40.6 & 9.8 & 8.7 & 27.7 \\
 & 1-shot & 37.3 & 45.3 & 9.9 & 30.2 & 37.7 & 44.7 & 9.9 & 30.4 \\
 & 3-shot & 38.8 & 48.3 & 12.3 & 31.8 & 39.8 & 47.8 & 14.1 & 32.6 \\
 & 5-shot & 33.4 & 34.7 & 11.6 & 26.6 & 34.8 & 34.7 & 12.2 & 27.6 \\
 \midrule
\multirow{4}{*}{Mistral-7B} & 0-shot & 30.5 & 0.0 & 3.8 & 19.5 & 30.8 & 0.0 & 5.0 & 19.9 \\
 & 1-shot & 33.2 & 0.4 & 5.3 & 21.4 & 34.5 & 0.4 & 6.0 & 22.4 \\
 & 3-shot & 36.2 & 5.6 & 7.6 & 24.2 & 36.6 & 6.6 & 8.2 & 24.7 \\
 & 5-shot & 36.6 & 7.0 & 8.2 & 24.8 & 37.2 & 7.5 & 8.8 & 25.3 \\
 \midrule
\multirow{4}{*}{Qwen-2.5-7B} & 0-shot & 34.6 & 0.0 & 6.1 & 22.4 & 34.9 & 0.0 & 6.4 & 22.6 \\
 & 1-shot & 34.0 & 0.0 & 5.3 & 21.9 & 35.0 & 0.1 & 6.2 & 22.7 \\
 & 3-shot & 32.6 & 28.7 & 4.8 & 24.6 & 33.4 & 25.4 & 6.2 & 24.8 \\
 & 5-shot & 32.4 & 21.0 & 5.1 & 23.5 & 33.5 & 19.6 & 5.1 & 24.0 \\
\bottomrule
\end{NiceTabular}
}
\caption{Full ablation results of zero- and few-shot inference.}
\label{table:ablation_num_shot}
\end{table*}

\begin{table*}[t]
\centering
\footnotesize
\resizebox{1\linewidth}{!}{
\begin{NiceTabular}{l|l|l|cccc|cccc}
\toprule
\multirow{2}{*}{Method} & \multirow{2}{*}{Unit} & \multirow{2}{*}{Number} & \multicolumn{4}{c}{Test Set 1 (Seen User/Unseen QA)} & \multicolumn{4}{c}{Test Set 2 (Unseen User/Unseen QA)} \\
& & & Persona & Schedule & Multi-Info & Total & Persona & Schedule & Multi-Info & Total \\
\bottomrule\toprule
\multirow{6}{*}{BM25} & Dialogue & 1 & 30.9 & 85.3 & 6.4 & 30.8 & 32.1 & 86.4 & 7.5 & 31.8 \\
 & Dialogue & 3 & 32.7 & 83.8 & 4.8 & 31.5 & 33.2 & 81.8 & 5.2 & 31.6 \\
 & Dialogue & 5 & 30.8 & 78.4 & 4.6 & 29.6 & 32.3 & 79.1 & 4.7 & 30.7 \\
 & Utterance & 5 & 34.1 & 78.5 & 6.1 & 31.9 & 35.0 & 78.0 & 7.0 & 32.5 \\
 & Utterance & 15 & 30.7 & 61.9 & 4.2 & 27.5 & 31.7 & 62.1 & 5.4 & 28.2 \\
 & Utterance & 25 & 30.3 & 38.0 & 4.2 & 24.2 & 30.4 & 38.7 & 4.6 & 24.4 \\
 \midrule
\multirow{6}{*}{Contriever} & Dialogue & 1 & 40.0 & 38.0 & 11.2 & 31.1 & 40.1 & 36.8 & 11.2 & 31.0 \\
 & Dialogue & 3 & 32.2 & 66.9 & 7.2 & 29.4 & 33.6 & 66.0 & 6.3 & 30.0 \\
 & Dialogue & 5 & 30.6 & 74.0 & 6.0 & 29.2 & 31.9 & 73.8 & 5.3 & 29.9 \\
 & Utterance & 5 & 42.6 & 70.3 & 9.8 & 36.6 & 43.1 & 71.8 & 9.1 & 37.1 \\
 & Utterance & 15 & 33.1 & 59.4 & 4.8 & 28.7 & 33.5 & 60.3 & 5.3 & 29.1 \\
 & Utterance & 25 & 30.8 & 34.5 & 4.3 & 24.1 & 31.3 & 35.8 & 5.0 & 24.7 \\
\bottomrule
\end{NiceTabular}
}
\caption{Full ablation results of various settings of Retrieval-augmented Generation.}
\label{table:ablation_rag_setting}
\end{table*}

\begin{table*}[t]
\centering
\footnotesize
\resizebox{1\linewidth}{!}{
\begin{NiceTabular}{l|l|cccc|cccc}
\toprule
\multirow{2}{*}{Hyperparam.} & \multirow{2}{*}{Value} & \multicolumn{4}{c}{Test Set 1 (Seen User/Unseen QA)} & \multicolumn{4}{c}{Test Set 2 (Unseen User/Unseen QA)} \\
& & Persona & Schedule & Multi-Info & Total & Persona & Schedule & Multi-Info & Total \\
\bottomrule\toprule
\multirow{6}{*}{Learning Rate} & 1e-6 & 16.1 & 19.6 & 3.4 & 12.9 & 16.9 & 18.2 & 3.8 & 13.3 \\
 & 3e-6 & 22.6 & 65.9 & 5.0 & 23.0 & 22.6 & 67.3 & 5.1 & 23.2 \\
 & 1e-5 & 30.0 & 67.0 & 10.1 & 28.4 & 29.3 & 66.0 & 10.0 & 27.8 \\
 & 3e-5 & 33.2 & 84.4 & 12.3 & 32.9 & 34.1 & 85.0 & 10.8 & 33.3 \\
 & 1e-4 & 36.5 & 87.5 & 15.8 & 35.7 & 34.8 & 87.5 & 14.4 & 34.5 \\
 & 3e-4 & 34.3 & 88.3 & 12.9 & 34.1 & 32.9 & 88.6 & 13.8 & 33.3 \\
 \midrule
\multirow{6}{*}{Rank} & 8 & 31.7 & 70.6 & 10.8 & 30.0 & 31.1 & 68.9 & 9.9 & 29.3 \\
 & 16 & 32.8 & 69.1 & 11.4 & 30.5 & 32.3 & 69.8 & 11.3 & 30.4 \\
 & 32 & 33.3 & 82.6 & 12.4 & 32.7 & 33.8 & 81.6 & 11.2 & 32.7 \\
 & 64 & 34.0 & 85.0 & 13.4 & 33.6 & 34.7 & 84.4 & 13.0 & 33.9 \\
 & 128 & 36.7 & 88.4 & 16.0 & 36.0 & 35.3 & 88.2 & 13.5 & 34.8 \\
 & 256 & 36.5 & 87.5 & 15.8 & 35.7 & 34.8 & 87.5 & 14.4 & 34.5 \\
\bottomrule
\end{NiceTabular}
}
\caption{Ablation on hyperparameters in training Llama-3.1 model with LoRA-based Supervised Fine-tuning.} 
\label{table:ablation_sft_hyperparam}
\end{table*}

\begin{table*}[t]
\centering
\footnotesize
\resizebox{1\linewidth}{!}{
\begin{NiceTabular}{l|l|cccc|cccc}
\toprule
\multirow{2}{*}{Hyperparam.} & \multirow{2}{*}{Value} & \multicolumn{4}{c}{Test Set 1 (Seen User/Unseen QA)} & \multicolumn{4}{c}{Test Set 2 (Unseen User/Unseen QA)} \\
& & Persona & Schedule & Multi-Info & Total & Persona & Schedule & Multi-Info & Total \\
\bottomrule\toprule
\multirow{6}{*}{Learning Rate} & 1e-6 & 28.7 & 10.1 & 5.3 & 19.9 & 29.1 & 11.1 & 6.1 & 20.4 \\
 & 3e-6 & 6.9 & 31.6 & 1.0 & 8.4 & 7.1 & 32.0 & 1.0 & 8.6 \\
 & 1e-5 & 7.2 & 27.3 & 0.7 & 8.0 & 6.8 & 30.2 & 1.0 & 8.2 \\
 & 3e-5 & 5.9 & 22.7 & 0.3 & 6.6 & 6.0 & 24.2 & 0.4 & 6.8 \\
 & 1e-4 & 8.7 & 3.0 & 3.0 & 6.2 & 8.4 & 3.2 & 3.2 & 6.0 \\
 & 3e-4 & 0.0 & 0.0 & 0.0 & 0.0 & 0.0 & 0.0 & 0.0 & 0.0 \\
 \midrule
\multirow{6}{*}{Rank} & 8 & 30.6 & 2.0 & 3.8 & 19.8 & 31.1 & 2.1 & 4.7 & 20.3 \\
 & 16 & 30.7 & 2.1 & 4.2 & 20 & 31.4 & 1.8 & 4.7 & 20.5 \\
 & 32 & 31.8 & 2.2 & 4.4 & 20.7 & 33.1 & 2.5 & 5.4 & 21.7 \\
 & 64 & 33.4 & 3.3 & 5.2 & 21.9 & 34.4 & 3.3 & 5.8 & 22.7 \\
 & 128 & 34.8 & 4.2 & 6.4 & 23.1 & 35.1 & 5.1 & 6.3 & 23.4 \\
 & 256 & 28.7 & 10.1 & 5.3 & 19.9 & 29.1 & 11.1 & 6.1 & 20.4 \\
\bottomrule
\end{NiceTabular}
}
\caption{Ablation on hyperparameters in training Llama-3.1 model with LoRA-based Direct Preference Optimization.}
\label{table:ablation_dpo_hyperparam}
\end{table*}

\begin{table*}[t]
\centering
\footnotesize
\resizebox{1\linewidth}{!}{
\begin{NiceTabular}{l|l|cccc|cccc}
\toprule
\multirow{2}{*}{Hyperparam.} & \multirow{2}{*}{Value} & \multicolumn{4}{c}{Test Set 1 (Seen User/Unseen QA)} & \multicolumn{4}{c}{Test Set 2 (Unseen User/Unseen QA)} \\
& & Persona & Schedule & Multi-Info & Total & Persona & Schedule & Multi-Info & Total \\
\bottomrule\toprule
\multirow{6}{*}{Learning Rate} & 1e-6 & 41.2 & 96.1 & 23.4 & 40.7 & 40.9 & 96.0   & 21.2 & 40.2 \\
 & 3e-6 & 48.1 & 98.1 & 18.4 & 44.6 & 47   & 98.1 & 22.4 & 44.4 \\
 & 1e-5 & 44.8 & 99.8 & 6.6  & 41.3 & 44.2 & 99.8 & 6.7  & 41.0   \\
 & 3e-5 & 49.1 & 98.7 & 8.1  & 44   & 47.6 & 98.8 & 7.8  & 43.1 \\
 & 1e-4 & 4.9  & 23.4 & 5.2  & 6.6  & 5.3  & 23.5 & 4.3  & 6.8  \\
 & 3e-4 & 2.5  & 0.0    & 1.1  & 1.7  & 2.4  & 0.0    & 1.4  & 1.7 \\
 \midrule
\multirow{6}{*}{Rank}  & 8.0   & 37.2 & 91.6 & 15.4 & 36.6 & 37   & 91.4 & 16.0   & 36.6 \\
 & 16  & 39.9 & 93.8 & 18.3 & 38.9 & 38.9 & 93.2 & 17.1 & 38.1 \\
 & 32  & 42.6 & 95.2 & 21.7 & 41.2 & 41.2 & 95.8 & 20.6 & 40.3 \\
 & 64  & 45.0   & 96.2 & 26.0   & 43.4 & 43.7 & 96.2 & 25.7 & 42.5 \\
 & 128 & 47.5 & 97.5 & 24.5 & 44.9 & 46.2 & 97.3 & 26.7 & 44.4 \\
 & 256 & 48.1 & 98.1 & 18.4 & 44.6 & 47.0   & 98.1 & 22.4 & 44.4 \\
\bottomrule
\end{NiceTabular}
}
\caption{Ablation on hyperparameters in training with LoRA-based Direct Preference Optimization done after Supervised Fine-Tuning (SFT$+$DPO).}
\label{table:ablation_sftdpo_hyperparam}
\end{table*}

\begin{figure*}[t]
  \centering
  \includegraphics[width=\linewidth]{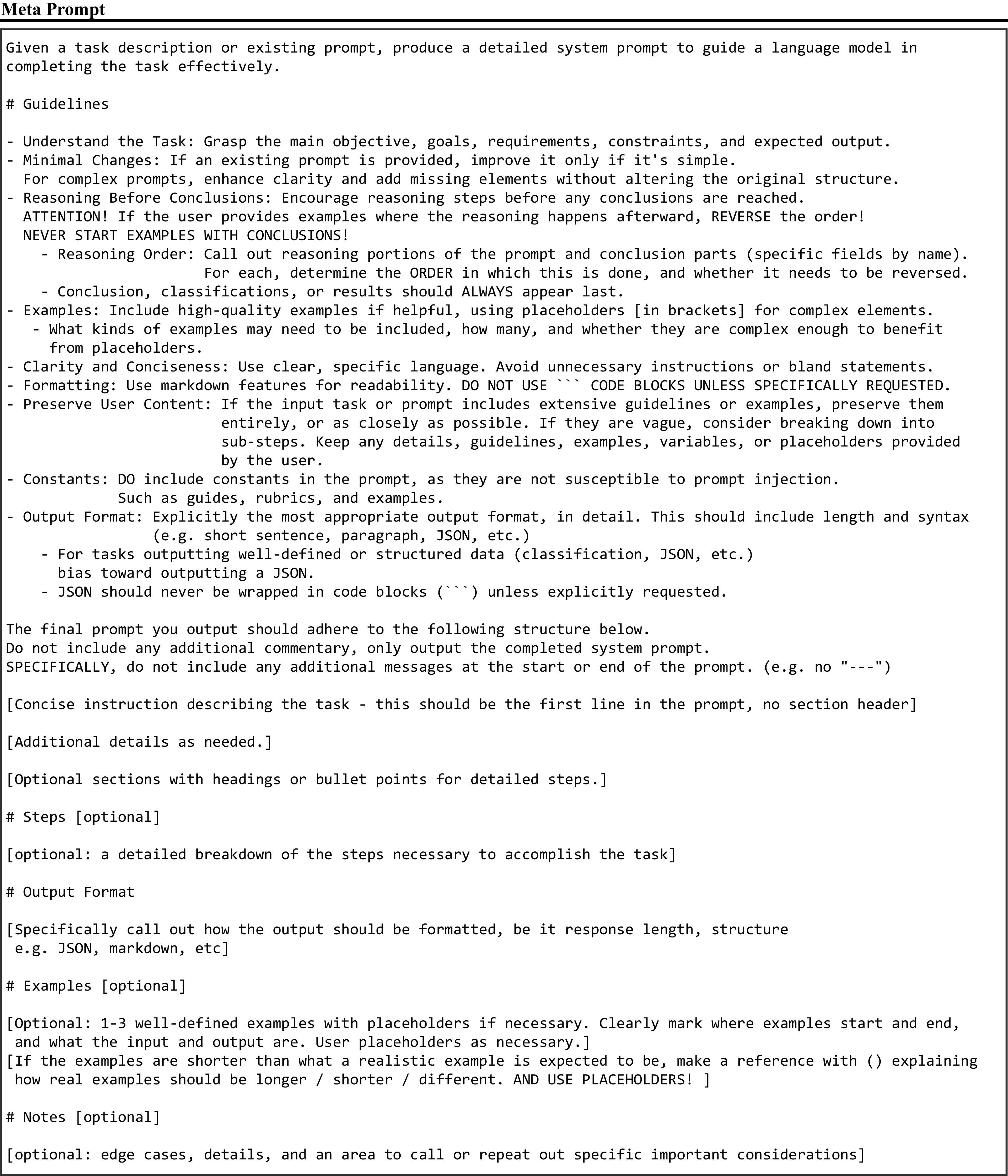}\\[-0.3em]
  \caption{\label{fig:meta_p} Meta-prompt for prompt optimization. The meta-prompt is used to generate the optimal prompt by prompting GPT-4o along with the task description. This meta-prompt is taken directly from OpenAI documentation and was created by OpenAI based on prompt engineering best practices and real-world experience (Source: \url{https://platform.openai.com/docs/guides/prompt-generation}). Generation setting: $\tau=0.0$ (greedy).}
\end{figure*}

\begin{figure*}[t]
  \centering
  \includegraphics[width=\linewidth]{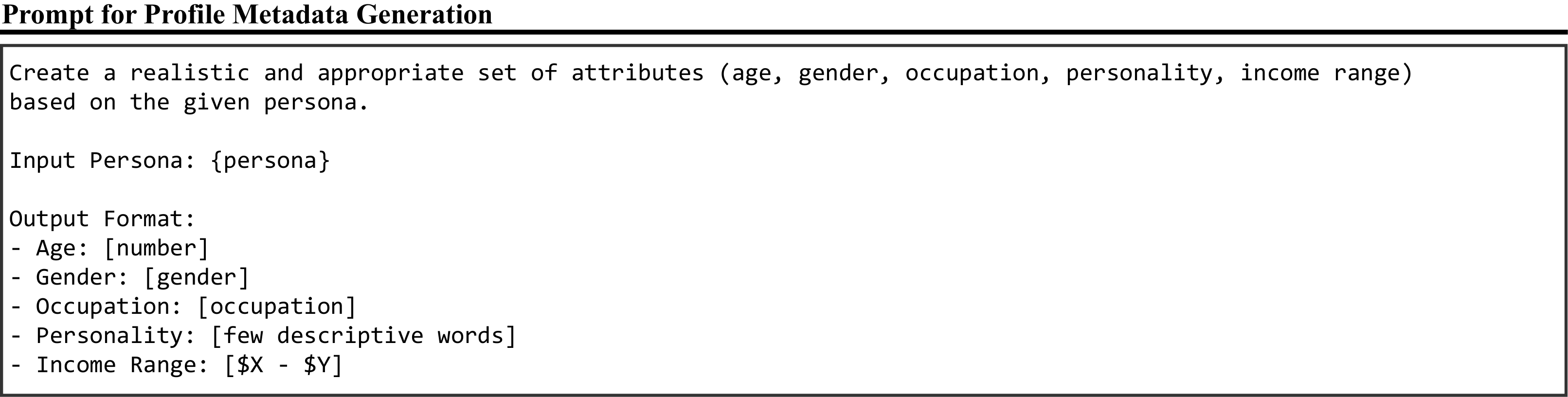}\\[-0.3em]
  \caption{\label{fig:p_profile_metadata} Prompt template for generating \textbf{profile metadata} in~\ours based on PersonaHub. The `Input Persona' is the description for one of the 1,500 randomly-sampled individuals from PersonaHub. Generation setting: $\tau=0.0$ (greedy).}
\end{figure*}

\begin{figure*}[t]
  \centering
  \includegraphics[width=\linewidth]{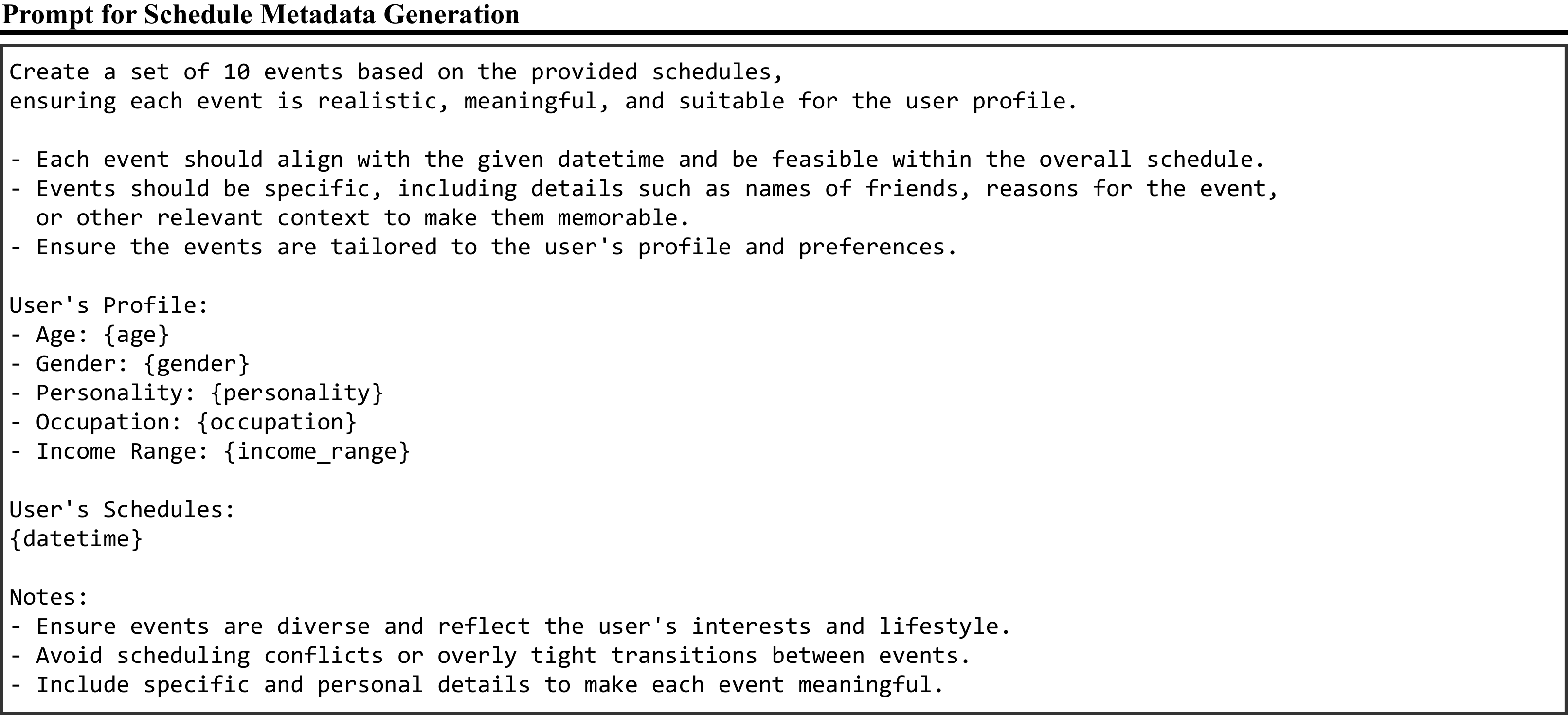}\\[-0.3em]
  \caption{\label{fig:p_schedule_metadata} Prompt template for generating \textbf{schedule metadata} in~\ours. The user's schedules are explicitly conditioned on the profile metadata from above to guarantee that they are realistic, meaningful, and suitable. For each user, we generate 10 different schedules. Generation setting: $\tau=0.0$ (greedy).}
\end{figure*}

\clearpage

% \subsection{Dialogue Generation}
% \subsection{QA Pair Generation}
% \subsection{2-hop Persona Pair Generation}
% \subsection{}
% \section{Example Appendix}
% \label{sec:appendix}

% This is an appendix.

\end{document}